%% file: main.tex
\documentclass{article}

\PassOptionsToPackage{numbers, compress}{natbib}

\usepackage[preprint]{neurips_data_2023}

\usepackage[utf8]{inputenc} 
\usepackage[T1]{fontenc}    
\usepackage{hyperref}       
\usepackage{url}            
\usepackage{booktabs}       
\usepackage{amsfonts}       
\usepackage{nicefrac}       
\usepackage{microtype}      
\usepackage{xcolor}         
\usepackage{amsbsy}
\usepackage{tabularx}
\usepackage{ragged2e}
\usepackage{graphicx}
\usepackage{multirow}
\usepackage{makecell}
\usepackage{subfigure}
\usepackage{caption}
\newcommand{\etal}{\textit{et al.}}

\newcommand{\eg}{\textit{e.g.}}
\newcommand{\ie}{\textit{i.e.}}

\title{Human Preference Score v2: A Solid Benchmark for Evaluating Human Preferences of\\ Text-to-Image Synthesis}

\author{
    Xiaoshi Wu$^{1*}$, 
    Yiming Hao$^{5*}$,
    Keqiang Sun$^{1}$, 
    Yixiong Chen$^{2}$, \\
    \textbf{
    Feng Zhu$^{3}$, 
    Rui Zhao$^{3,4}$, 
    Hongsheng Li$^{1, 5}$
    }
\vspace{0.5em}\\
    $^{1}$Multimedia Laboratory, The Chinese University of Hong Kong \quad $^{2}$CUHK-SZ, SRIBD\\
    $^{3}$SenseTime Research \quad $^{4}$Qing Yuan Research Institute, Shanghai Jiao Tong University \\
    $^{5}$Centre for Perceptual and Interactive Intelligence (CPII) \\ 
    \texttt{\small \{wuxiaoshi@link, kqsun@link, hsli@ee\}.cuhk.edu.hk} \\
    \texttt{\small ymhao@cpii.hk,~yixiongchen@link.cuhk.edu.cn}\\
    \texttt{\small \{zhufeng, zhaorui\}@sensetime.com}
}

\begin{document}

\maketitle

\thispagestyle{empty}
\footnotetext[1]{\noindent Equal contribution.}

\begin{abstract}
Recent text-to-image generative models can generate high-fidelity images from text inputs, but the quality of these generated images cannot be accurately evaluated by existing evaluation metrics. 
To address this issue, we introduce Human Preference Dataset v2 (HPD v2), a large-scale dataset that captures human preferences on images from a wide range of sources. 
HPD v2 comprises 798,090 human preference choices on 433,760 pairs of images, making it the largest dataset of its kind.
The text prompts and images are deliberately collected to eliminate potential bias, which is a common issue in previous datasets.
By fine-tuning CLIP on HPD v2, we obtain Human Preference Score v2 (HPS v2), a scoring model that can more accurately predict human preferences on generated images.
Our experiments demonstrate that HPS v2 generalizes better than previous metrics across various image distributions and is responsive to algorithmic improvements of text-to-image generative models, making it a preferable evaluation metric for these models.
We also investigate the design of the evaluation prompts for text-to-image generative models, to make the evaluation stable, fair and easy-to-use.
Finally, we establish a benchmark for text-to-image generative models using HPS v2, which includes a set of recent text-to-image models from the academic, community and industry.
The code and dataset is available at \href{https://github.com/tgxs002/HPSv2}{https://github.com/tgxs002/HPSv2}
\end{abstract}

\section{Introduction}
Recent advances in text-to-image synthesis~\cite{dalle, dalle2, stable_diffusion, glide, ding2022cogview2} have made it possible to generate high-fidelity images based on text prompts. 
However, images generated with different random seed often have varying quality, and previous works~\cite{wu2023better, xu2023imagereward,kirstain2023pick} demonstrate that popular metrics, such as Inception Score (IS) \cite{inceptionscore}, Fréchet Inception Distance (FID) \cite{fid}, and CLIP Score~\cite{clip}, do not correlate well with human preferences on these images. 
Therefore, human preference is an important but poorly tracked aspect of text-to-image generative models. 
To facilitate research in this area, we construct a large-scale dataset annotated with human preferences, namely Human Preference Dataset v2 (HPD v2).
We also established a benchmark based on a preference prediction model, Human Preference Score v2 (HPS v2), which is trained on HPD v2, to validate algorithmic developments for human-aligned image synthesis.

\begin{figure}
    \centering
    \includegraphics[width=0.8\linewidth]{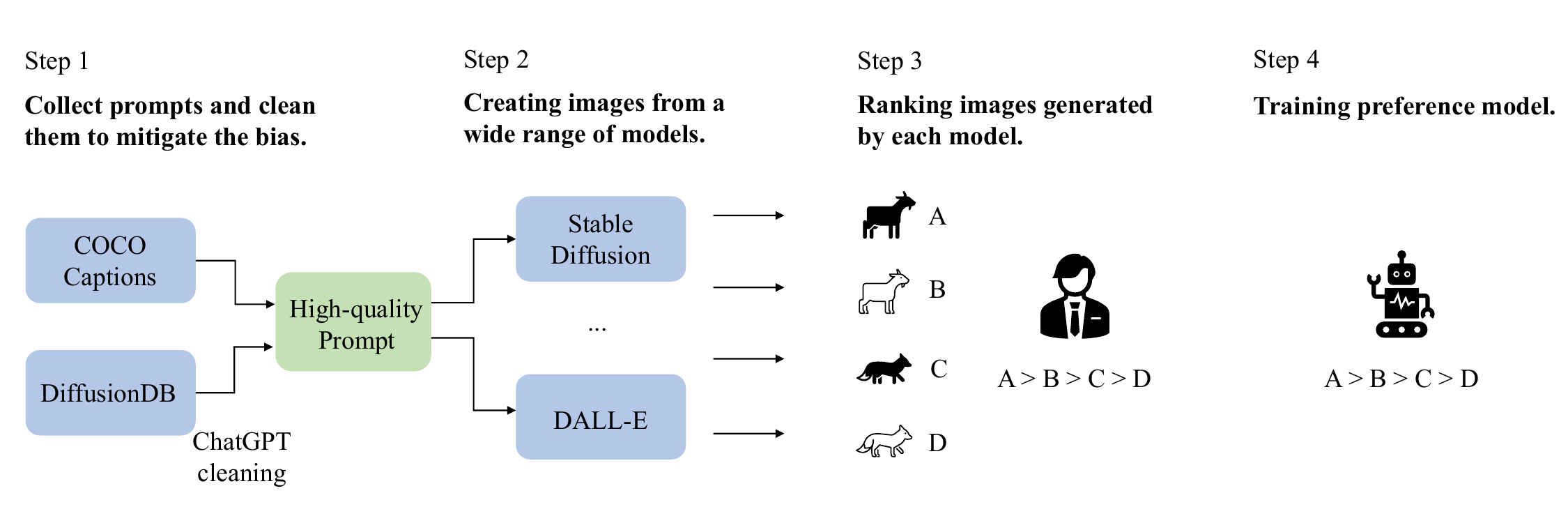}
    \caption{Overview of this work. We firstly collect Human Preference Dataset v2 (HPD v2) and then train a preference prediction model, Human Preference Score v2 (HPS v2), on it.}
    \label{fig:overview}
\end{figure}

HPD v2 is a comprehensive dataset of human preferences comparing images from various sources, including 9 text-to-image generative models and the COCO Captions dataset~\cite{cococaptions}. 
It comprises 798k human-annotated pairwise comparisons of images generated from the same prompt, making it the largest dataset of its kind. 
The dataset addresses issues of bias that occurs in previous datasets~\cite{wu2023better, xu2023imagereward, kirstain2023pick}.
The first type of bias derives from the image source.
Previous datasets~\cite{wu2023better, xu2023imagereward, kirstain2023pick} mainly contain images generated from Stable Diffusion~\cite{stable_diffusion} and its variants.
Therefore, we cannot verify whether the models trained and evaluated exclusively on these datasets can generalize to other image distributions.
HPD v2 incorporates images generated from 9 recent models~\cite{ding2022cogview2, glide, vqgan, liu2021fusedream, zhou2021lafite, stable_diffusion, gu2022vector, dalle}, as well as real images from the COCO Captions dataset~\cite{cococaptions}, enabling better evaluation on the generalization capability.
Another source of bias is in text prompts.
User-written prompts, such as prompts in DiffusionDB~\cite{wang2022diffusiondb}, often follow a specific organization of description plus several style words, where the style words often contain contradictions, making it harder for annotators to understand. (see Tab.~\ref{tab:clean_prompt} for examples)
The style words are also highly biased, leading to issues in training and evaluation.
To tackle this bias, we employ ChatGPT to remove style words and organize prompts into a clearly written sentence.

Based on HPD v2, a preference prediction model, Human Preference Score v2 (HPS v2), is trained to score generated images based on its likelihood to be preferred by humans.
Experiments show that HPS v2 generalizes better than previous models, including HPS v1~\cite{wu2023better}, ImageReward~\cite{xu2023imagereward} and PickScore~\cite{kirstain2023pick}.

We further study how to evaluate text-to-image generative models using HPS v2, and set up a benchmark for them.
This is mainly about choosing a proper set of prompts.
Pick-a-Pic~\cite{kirstain2023pick} proposes to use prompt in its dataset. However, they did not study which set of prompts to choose, and how many prompts need to be included.
DrawBench is a list of prompts introduced in Imagen~\cite{imagen}, but we find that it is not large enough for stable evaluation.
DiffusionDB~\cite{wang2022diffusiondb} is a large scale database of user-written prompts and images generated from them.
However, a large portion of these prompts are biased towards certain dataset-specific style words, and is not fair for models trained with other data.
To mitigate this issue, we use ChatGPT to clean prompts from DiffusionDB~\cite{wang2022diffusiondb}.
We carefully study the number of prompts for evaluation to ensure that the average score is statistically stable.
Finally, we obtain 4 lists of prompts for evaluation and benchmarking, which are categorized by styles: ``Animation'', ``Concept-art'', ``Painting'', and ``Photo''.

We also show that HPS v2 is sensitive to algorithmic improvements.
Firstly, we evaluate the effectiveness of a straightforward test-time trick of blending the input noise for a diffusion model with a prior image.
We also conduct experiments to validate the quantitative improvements of a method proposed by Wu \etal~\cite{wu2023better}.
Our experiments show that HPS v2 is sensitive to algorithmic improvements, and can be used to evaluate text-to-image generative models.

Our contributions are as follows:
1) HPD v2, a large-scale, well-annotated dataset for researches of human preferences on images generated by text-to-image generative models.
2) HPS v2, a better human preference prediction model against existing ones, for which we show several example usages to demonstrate its sensitivity and accuracy.
3) A fair, stable and easy-to-use set of evaluation prompts for text-to-image generative models.

\section{Related Work}
\subsection{Text-to-image generative models}
Text-to-image generative models enable users to create images based on textual input. DALL·E~\cite{dalle} was the first to achieve high-quality open-domain text-to-image generation. Since then, various architectures and modeling have been explored to enhance image quality. These include autoregressive models~\cite{dalle, ding2022cogview2}, GANs~\cite{sauer2023stylegan}, and diffusion models~\cite{stable_diffusion, dalle2, imagen}. Among them, diffusion models have demonstrated superior computational efficiency and the ability to produce high-quality samples~\cite{dalle2}. However, despite the capability of current text-to-image diffusion models to generate high-fidelity images, they often fail to align well with human preferences. Recent researches in this field has attempted to improve the image quality of diffusion models~\cite{wu2023better, lee2023aligning}, but the evaluation of these improvements remains challenging due to the lack of appropriate metrics and benchmarks.  
Therefore, there is a need to develop new evaluation metrics and benchmarks to effectively validate algorithmic enhancements.

\subsection{Image quality evaluation for generative models}
Inception Score (IS)~\cite{inceptionscore} and Fréchet Inception Distance (FID) \cite{fid} are widely used in the evaluation of image generative models.
However, they do not correlate well with human preferences for images generated by recent text-to-image models~\cite{kirstain2023pick, wu2023better}.
LPIPS~\cite{zhang2018unreasonable} measures perceptual similarity between image pairs by fitting a pretrained model on a dataset of human perception on similarity.
Similar methodology is used in HYPE~\cite{zhou2019hype} for evaluating generated human faces.
Several recent works~\cite{wu2023better, xu2023imagereward, kirstain2023pick} propose to fine-tune visual-language models (VLM) on human choices on images generated with the same prompt, and use fine-tuned VLMs to serve as proxies for human evaluation, but these models are tuned exclusively on images from Stable Diffusion~\cite{stable_diffusion} and its variants, their generalization capabilities are not yet validated.

\subsection{Image quality dataset}
AVA~\cite{ava} is a dataset of human ratings given to photographs.
Similar to AVA, Simulacra~\cite{sac} collects human ratings, but instead for images generated from text.
Both AVA and Simulacra are datasets of image ratings, which are absolute scores for images.
They have been used to train the Aesthetic Score Predictor~\cite{schuhmann2022laion}, which participates in the data cleaning pipeline of Stable Diffusion~\cite{stable_diffusion}.
InstructGPT~\cite{ouyang2022training} proposes to align a pre-trained language model with human feedback.
The idea of Reinforcement Learning from Human Feedback (RLHF) is also explored for visual content creation~\cite{lee2023aligning, wu2023better}.
This requires comparison between images generated from the same prompt.
HPD v1~\cite{wu2023better} was collected under this context, and it improves the image quality by aligning text-to-image models with human preference.
Similar to HPD v1, ImageReward~\cite{xu2023imagereward} and Pick-a-Pic~\cite{kirstain2023pick} are datasets of comparison between paired images, focusing on evaluation of generated images.
AGIQA-1K~\cite{zhang2023perceptual} and AGIQA-3K~\cite{li2023agiqa3k} are collected similarly, but the dataset size is considerably smaller.
Our HPD v2 includes more images from a wider range of sources than existing works, enabling a more comprehensive evaluation of these preference prediction models.
A head-to-head comparison with other datasets is presented in Tab.~\ref{tab:dataset_comparison}.
Fake2M~\cite{lu2023seeing} is another related dataset that focuses on visual Turing test of distinguishing generated images.


\section{Human Preference Dataset v2}
Human Preference Dataset v2 (HPD v2) is a large-scale, cleanly-annotated dataset of human preferences for images generated from text prompts. 
In total, the dataset contains 798k pairs of binary preference choices for 434k images. 
Each pair contains two images generated by different models using the same prompt, and is annotated with a binary choice made by one annotator.
The collection pipeline of HPD v2 includes prompt collection (Sec.\ref{sec:prompt_collection}), image collection (Sec.\ref{sec:image_collection}), and preference annotation (Sec.\ref{sec:preference_annotation}). 
Detailed statistics about the dataset are presented in Sec.\ref{sec:statistics}.
A head-to-head comparison with previous datasets is made in Sec.~\ref{sec:dataset_comparison}.

\input{tables/prompt_cleaning}

\subsection{Prompt collection}
\label{sec:prompt_collection}

\noindent \textbf{Prompt sources.} A naive idea is to use captions from image-caption datasets, such as COCO Captions~\cite{cococaptions} and LAION~\cite{schuhmann2022laion}.
However, these captions are mostly for describing real images, which may not reflect the general interest of generative model users.
Many ``fiction'' content can only be created from the imagination of generative model users.

There are two recent sources containing user-written prompts.
DrawBench is a list of prompts proposed in Imagen~\cite{imagen} to evaluate text-to-image generative models.
It is a collection of 200 prompts organized into 11 categories, designed to test various aspects of a model's capabilities. 
However, the size only supports the purpose of evaluation.
DiffusionDB~\cite{wang2022diffusiondb} is a database containing a wide range of user-written prompts and generated images. 
However, a significant portion of the prompts in the database is biased towards certain styles.
For instance, around 15.0\% of the prompts in DiffusionDB include the name ``Greg Rutkowski'', 28.5\% include ``artstation''.
Note that the effectiveness of these words is highly dependent on the training data of generative models.
A model trained without images from artstation, or at least without ``artstation'' in its training prompts, may not respond to such style words.
It is unfair to test these models on the biased prompts.
There are also conflict contents in these prompts.
For example, consider the first prompt in Tab.\ref{tab:clean_prompt}, ``renaissance oil painting'' is conflict with ``realism'' and ``hyper realistic''. 
These conflicts may confuse annotators in evaluating image-text alignment.

We collect three groups of prompts for different purposes: training a preference prediction model, testing a preference prediction model, and benchmarking generative models.
All three groups consist of prompts from COCO Captions~\cite{cococaptions} and prompts from DiffusionDB~\cite{wang2022diffusiondb} cleaned by ChatGPT.
Detailed statistics will be introduced in Sec.~\ref{sec:statistics} and Sec.~\ref{sec:benchmark}.

\begin{figure}
\centering
\begin{minipage}{.49\textwidth}
  \centering
  \includegraphics[width=1.0\linewidth]{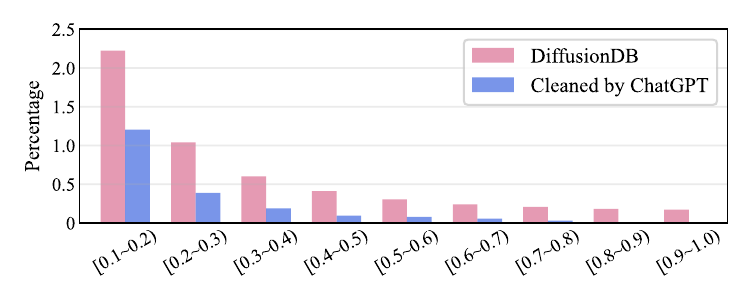}
  \caption{Histogram of NSFW score distribution. The first bucket counting ``safe'' prompts is omitted for better visualization.}
  \label{fig:nsfw_histogram}
\end{minipage}%
\hfill
\begin{minipage}{.49\textwidth}
  \centering
  \includegraphics[width=1.0\linewidth]{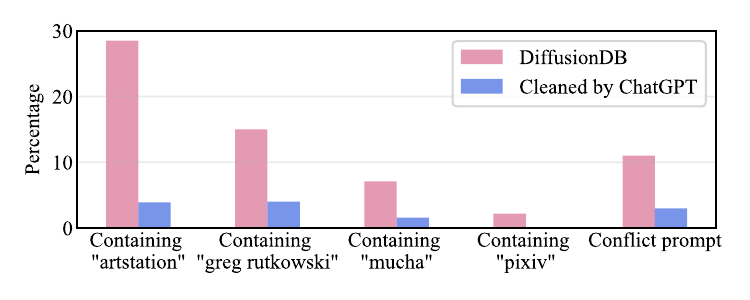}
  \caption{Frequencies of certain style words and conflict prompt. Confliction is judged by ChatGPT.}
  \label{fig:word_stat}
\end{minipage}
\end{figure}

\noindent \textbf{Prompt cleaning using ChatGPT.} To address the above issues, we utilize ChatGPT to rewrite prompts in DiffusionDB~\cite{wang2022diffusiondb}. 
The instruction for ChatGPT is elaborated in Appendix~\ref{sec:appendix_cleaning}.
Prompts for image generation and benchmarking generative models are all cleaned with the same pipeline.
In Tab.~\ref{tab:clean_prompt}, we show examples of prompts before and after cleaning.
The output prompts are clearly written in one sentence with less style words, which are easier for our annotators to understand.
Fig.~\ref{fig:nsfw_histogram} shows the histogram of NSFW score~\cite{Detoxify} distribution before and after cleaned by ChatGPT.
The effect is significant for buckets of high NSFW score.
Fig.~\ref{fig:word_stat} shows frequencies of certain style words and conflict prompt, where confliction is judged by ChatGPT.
The frequency of platform names and artist names decreases, and there are less conflict prompts after cleaning.

\begin{table}[ht]
    \caption{Image sources of HPD v2.}
    \label{tab:image_source}
    \centering
    \resizebox{0.65\linewidth}{!}{
        \input{tables/source_model}
    }
\end{table}

\subsection{Image collection}
\label{sec:image_collection}

In order to train and evaluate a preference prediction model, we generate images using different models and the same prompt for human preference annotation. 
For each prompt, we generate several images using different models.
We also add the real image for prompts in COCO Captions.
Tab.~\ref{tab:image_source} summarizes image sources of HPD v2. 
The dataset comprises images generated by models of different architectures, scales, resulting in a high degree of diversity. 
This diversity allows for a comprehensive evaluation of a preference prediction model's generalization capability and facilitates the training of a more generalizable model. 
The training set contains images from 4 models + COCO images, while the test set contains 9 models + COCO images.
The 5 unseen models enable us to validate the generalization capability of the trained preference prediction model.
For information on the inference parameters of each model, please refer to Appendix~\ref{sec:generation_details}.

\subsection{Preference annotation}
\label{sec:preference_annotation}

To gather preference rankings for a given prompt and its corresponding generated images, we employ annotators to assess and rank the images based on their preference. The preference may be influenced by factors such as image-text alignment and visual attractiveness. While these two factors can sometimes conflict with each other, understanding the trade-off is crucial.

\begin{table}[ht]
    \hspace{0.7cm}
    \begin{minipage}{.4\linewidth}
        \vspace{-1cm}
        \caption{Annotator statistics.}
        \vspace{0.2cm}
        \label{tab:annotator}
        \centering
        \resizebox{1.0\linewidth}{!}{
            \input{tables/annotator}
        }
    \end{minipage}
    \hfill
    \begin{minipage}{.4\textwidth}
      \centering
        \captionof{figure}{Distribution of inter-annotator agreement. The agreement is computed on pairwise comparison between images in HPD v2 test set.}
      \includegraphics[width=1.0\linewidth]{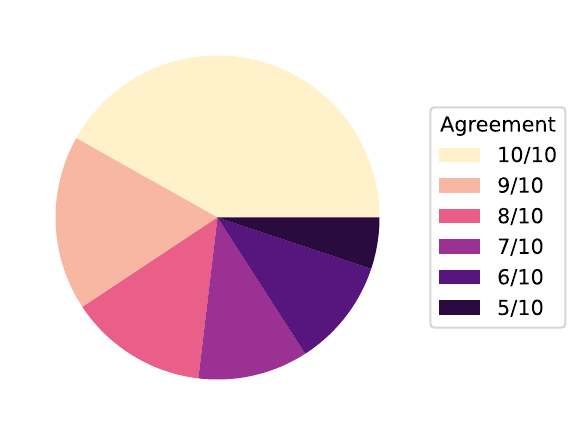}
      \label{fig:agreement_histogram}
    \end{minipage}
    \hspace{0.7cm}
\end{table}

\input{tables/human_vs_human}

\begin{figure}
    \centering
    \caption{Inter-annotator agreement between models plotted against absolute difference in HPS v2.}
    \includegraphics[width=0.7\linewidth]{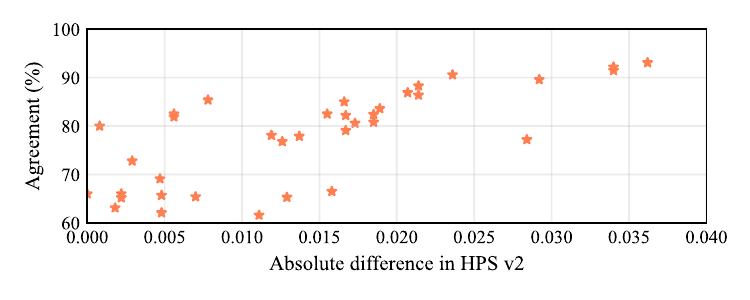}
    \label{fig:agreement_hps}
\end{figure}

We have assembled a team of 57 contractors, consisting of 50 labelers and 7 quality control checkers, to annotate our data. Each labeler undergoes an orientation to familiarize themselves with our requirements and the annotation system. Detailed guidelines regarding our requirements are documented in Appendix~\ref{sec:appendix_annotation}. Following the orientation, labelers are required to complete a test, which involves a task of ranking 300 groups of images. The checkers evaluate the annotation quality based on these tests. At least 10\% of the tasks are cross-checked by the checkers, who examine the top-ranked choice made by the labeler and calculate the recall @ 2 compared to the checkers' answer.

As preference is subjective, there is no fixed standard for acceptance. Decisions are made on a case-by-case basis. Even if a labeler has low recall during the test, they may still be hired if they can adequately explain their decisions. The overall hiring rate is 61\%. 
Tab.~\ref{tab:annotator} presents statistical information about our annotators. It is worth noting that our annotators do not possess specialized training in art appreciation. They are a group of average annotators rather than experts in the field. Spot tests continue to be conducted after the labeler is hired to ensure ongoing quality.

In Tab.~\ref{tab:huamn_vs_human}, we present the inter-labeler agreement for preferences on each pair of models in our test set. 
As depicted in Figure~\ref{fig:agreement_hps}, we observe that the agreement between models with similar quality is relatively lower, whereas the agreement is higher for comparisons between models with substantial disparities.

\subsection{Statistics}
\label{sec:statistics}
HPD v2 comprises a test split and a training split. 
The test split consists of 400 groups of images.
Among them, 300 groups use prompts from DiffusionDB~\cite{wang2022diffusiondb}, and 100 groups use prompts from COCO Captions~\cite{cococaptions}.
Each group contains 9 images generated from a text-to-image generative model.
For the 100 groups using prompts from COCO Captions~\cite{cococaptions}, we add one more real image from the COCO Captions dataset~\cite{cococaptions}.
Each group in the test set is annotated by 10 distinct annotators to ensure more stable evaluation and to enable the study of diversity in human preference, resulting in 153,000 binary comparisons.
The training split of HPD v2 contains 107,515 groups of images, each containing 4 images from 4 models (or real image), ranked by one annotator, which corresponds to 645,090 human binary comparisons.
28,172 prompts in the training set are from COCO Captions~\cite{cococaptions}, while others are from DiffusionDB~\cite{wang2022diffusiondb}.

\input{tables/dataset_comparison}

\subsection{Comparison with related datasets}
\label{sec:dataset_comparison}
A head-to-head statistical comparison with HPD v1~\cite{wu2023better}, ImageReward~\cite{xu2023imagereward}, and Pick-a-Pic~\cite{kirstain2023pick} is presented in Tab.~\ref{tab:dataset_comparison}.
HPD v1 and Pick-a-Pic collect preference data from the internet, while ImageReward and HPD v2 are collected from contract workers. 
Although it is easier to distribute a survey via online platforms like the Discord channel or web applications, the anonymous nature of the internet poses difficulty for quality control and analysis on potential bias.
For example, 34.9\% of instances in Pick-a-Pic has a prompt that repeats for more than 50 times, these repeated prompts only take up 4.7\% of total unique prompts.
On the contrary, we have detailed statistics about the background of our annotators, which guarantees a better transparency for our dataset. 
Prompts from HPD v1 and ImageReward are collected from Discord channels, which show a strong bias towards certain styles.
The bias problem is less severe in HPD v2.

\section{Human Preference Score v2}
In this section, we introduce the training of Human Preference Score v2 (HPS v2), and present a comprehensive evaluation of existing preference prediction models on the test set of HPD v2 and ImageReward~\cite{xu2023imagereward}.

\subsection{Training}
CLIP~\cite{clip} is a base-model that aligns image and text to the same embedding space.
It has an image encoder to encode an image into a visual feature, and a text encoder to encode a caption into a text feature.
The cosine similarity between the visual and text features measures the alignment between the input image and caption.
However, CLIP~\cite{clip} does not work well for zero-shot human preference prediction, as shown in Tab.~\ref{tab:evaluation}.
For more background information about the CLIP model, please refer to Appendix~\ref{sec:ablations}.

HPS v2 is trained by fine-tuning CLIP~\cite{clip} on HPD v2.
Each instance in the training set contains a pair of images $\{x_1, x_2\}$ with prompt $p$, which is labeled with $y=\left[1, 0\right]$ if image $x_1$ is preferred over $x_2$, otherwise  $y=\left[0, 1\right]$. 
The CLIP model can be viewed as a score function $s$ that computes the similarity between prompt $p$ and image $x$:
\begin{equation}
\label{eq:sim}
    s_\theta(p, x) = \frac{\mathrm{Enc}_{\rm txt}(p) \cdot \mathrm{Enc}_{\rm img}(x)}{\tau},
\end{equation}
where $\tau$ is the learned temperature scalar of the CLIP model, and $\theta$ is the parameters in CLIP.
The predicted preference $\hat{y}_i$ is calculated as:
\begin{equation}
    \hat{y_i} = \frac{\exp(s_\theta(p, x_i))}{\sum_{j=1}^2 \exp(s_\theta(p, x_j))}.
\end{equation}

$\theta$ is optimized by minimizing the KL-divergence:
\begin{equation}
    L_{\rm pref} = \sum_{j=1}^2 y_i(\log y_i - \log\hat{y_j}).
\end{equation}
We fine-tune ViT-H/14 trained by OpenCLIP ~\cite{clip, openclip} by optimizing $L_{\rm pref}$ on HPD v2 for 4,000 steps with the AdamW optimizer~\cite{adam, adamw}, a learning rate of $3.3\times10^{-6}$, a weight decay of 0.35, a batch size of 128 and a warm-up period of 500 steps, following a cosine learning rate schedule.
When fine-tuning a pre-trained model on limited data, a common practice is to freeze or decrease the learning rate of the first few layers~\cite{he2022masked, vasconcelos2022proper, Zhai_2022_CVPR}.
We train the last 20 layers of the CLIP image encoder and the last 11 layers of the CLIP text encoder. 
Hyper-parameters are determined by Bayesian optimization with the target of accuracy on the HPD v2 test set.
\label{sec:Train}

\input{tables/evaluation}

\subsection{Experiments}
\label{sec:model_eval}
In Tab.~\ref{tab:evaluation}, we report the accuracy of the baseline models on test sets of ImageReward~\cite{xu2023imagereward} and HPD v2, the consistency between human annotators, and the performance of HPS v2.
Different from the test set of ImageReward~\cite{xu2023imagereward} that only contains images generated by Stable Diffusion, the images of HPD v2 test set cover a much wider range of image distribution, as demonstrated in Tab.~\ref{tab:image_source}, and can more comprehensively evaluate a preference prediction model's generalization capability.
HPS v2 exhibits better accuracy on both benchmarks, demonstrating its strong capability of generalization.
In Appendix~\ref{sec:pairwise_acc}, we report the decomposed accuracy between models.
In Appendix~\ref{sec:ablations}, we report ablation experiments on hyper-parameter choices.

It should be noted that although human preferences are generally diverse, the average preference of multiple humans can reflect humans' general tendency.
Since HPS v2 is trained on preferences from many annotators, it is able to learn humans' average preferences, which can exceed a single person's accuracy (Single Human vs. Single Human).
Besides a single person's accuracy, we also report the accuracy of a single person compared to an averaged ranking, which is still higher than HPS v2.

\section{Benchmarking Text-to-Image Generative Models}
\label{sec:protocol}

\input{tables/benchmark}
\label{sec:benchmark}
Similar to HPS v1, HPS v2 evaluates the preference score of a generated image $x$ given a reference prompt $p$ as follows:
\begin{equation}
    s_\theta(p, x) = \mathrm{Enc}_{\rm txt}(p) \cdot \mathrm{Enc}_{\rm img}(x) \times 100,
\end{equation}
where the notation is the same as Equation~\ref{eq:sim}.
We evaluate models separately for different styles using prompts obtained from ChatGPT, as described in Sec. \ref{sec:prompt_collection}. 
To this end, we use a set of evaluation prompts that involves testing a model on a total of 3200 prompts, with 800 prompts for each of the following styles: ``Animation'', ``Concept-art'', ``Painting'', and ``Photo''. 
The first 3 styles are popular in DiffusionDB~\cite{wang2022diffusiondb} based on our observation.
For each style, 800 prompts are divided into groups of 80, and evaluated separately. We report the mean and standard deviation of 10 groups. The size of 800 prompts ensures that HPS v2 is statistically stable across all evaluated models, while also avoiding excessive computational overhead. 

Tab.~\ref{tab:benchmark} shows a benchmark of recent text-to-image generative models from the academic, community and industry.
It's an interesting trend that popular models from the community are consistently out-performing models from the academic.
We also plotted HPS v2 against existing evaluation metrics, CLIP Score~\cite{clip} and Aesthetic Score~\cite{schuhmann2022laion}. 
An interesting data point is CogView2~\cite{ding2022cogview2}.
It generates images with high aesthetic quality, but has worse controllability.
Inference details of each model on the benchmark are elaborated in Appendix~\ref{sec:generation_details}.

In Appendix~\ref{sec:example_usages}, we show more example usages of HPS v2 other than benchmarking text-to-image models.

\begin{figure}[ht]
    \centering
    \caption{Correlation between HPS v2 and other models.}
    \hfill
    \begin{minipage}{0.47\textwidth}
        \centering
        \includegraphics[width=1.0\linewidth]{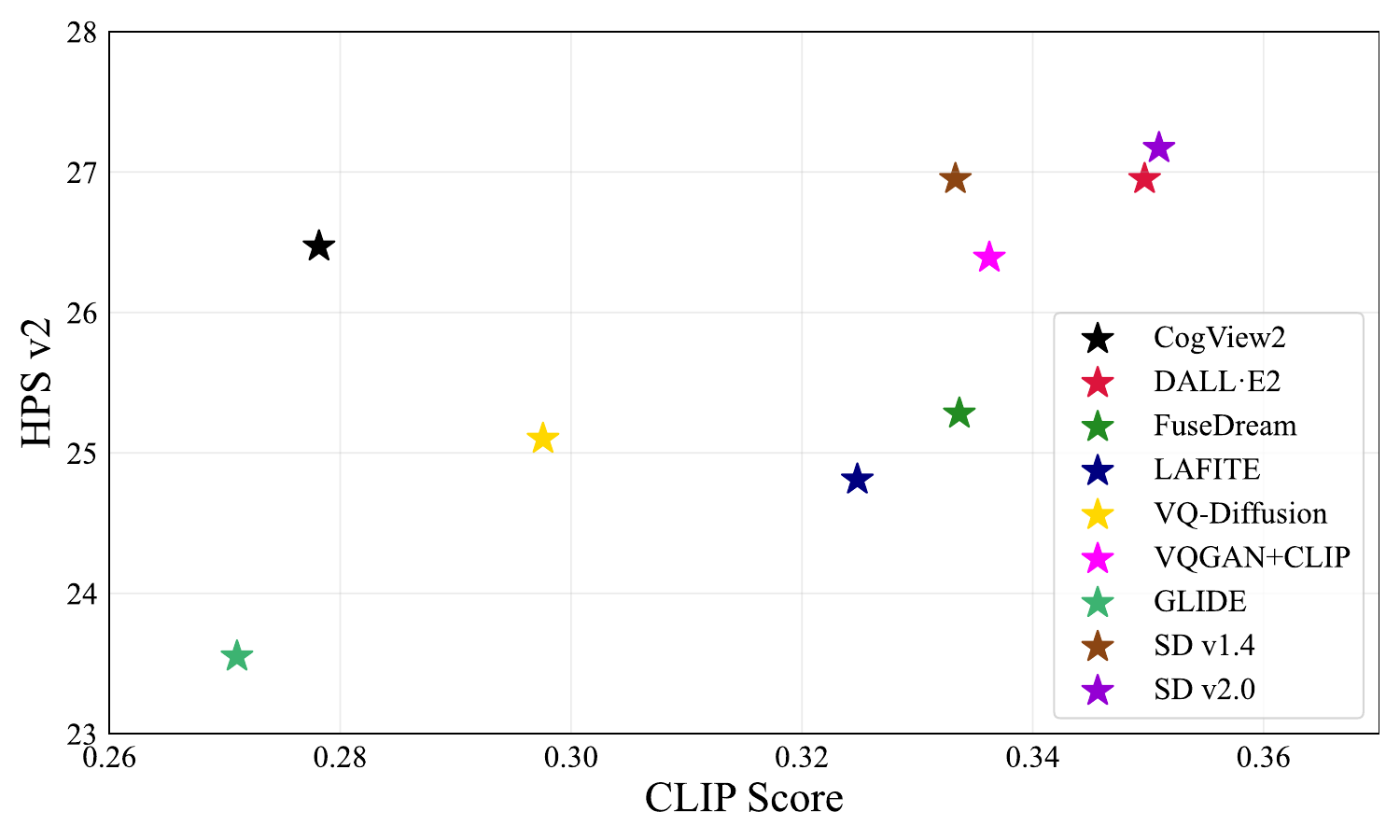}
        \label{fig:clip_hps}
    \end{minipage}
    \hfill
    \begin{minipage}{0.47\textwidth}
        \centering
        \includegraphics[width=1.0\linewidth]{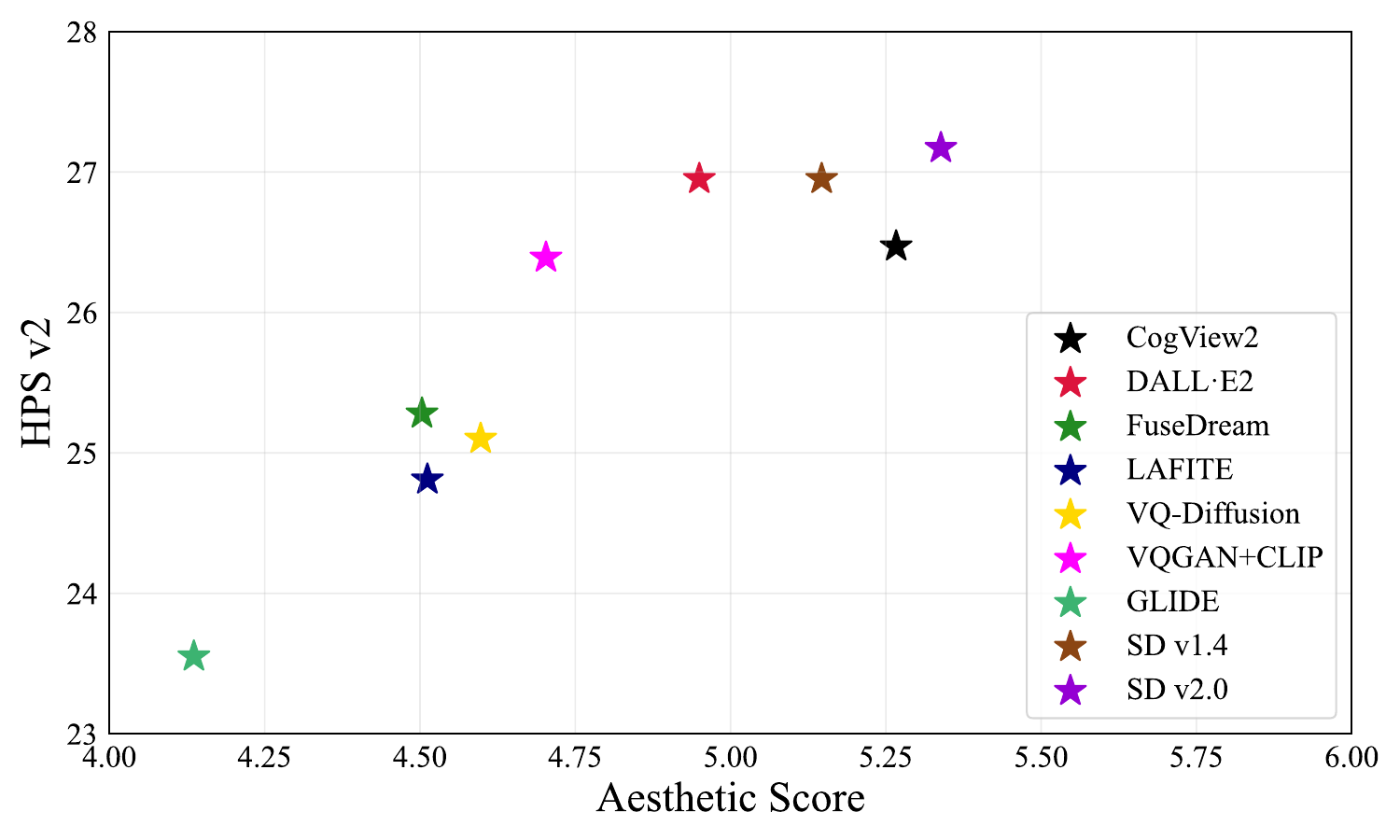}
        \label{fig:aesthetic_hps}
    \end{minipage}
\end{figure}

\section{Limitations}
There are several potential limitations about this work.
The prompts are sourced from DiffusionDB~\cite{wang2022diffusiondb} and COCO Captions~\cite{cococaptions}, and the choice of the 3 styles of our benchmark except ``Photo'' is summarized based on the general interest we observed in DiffusionDB.
Although DiffusionDB provides a massive collection of prompts that reflects community users' demands, there are indeed some overlooked topics, such as logo and graphic design.
These demands also play a crucial role in design industries and have unique criteria or preferences compared to general image creation. 
Using ChatGPT for prompt cleaning may also introduce potential bias, although currently we did not find such patterns.
Also, this dataset is annotated by 57 annotators, and thus it may suffer from bias of them.
Another limitation is that this work did not study the effect of image resolution on human preferences, though this can make a big impact as the resolution of generated images keeps increasing.

Generative models potentially have negative social impacts by their nature.
The advancement in generative models often means more plausible generated content, which may be maliciously utilized for the spread of more convincing misinformation and fake content.
Additionally, there is a risk of amplifying existing biases and stereotypes present in the training data.

\label{sec:limitations}

\section{Conclusion}

In this work, we design a less biased pipeline for collecting prompts and images for human preference annotations. 
With collected data, we present Human Preference Dataset v2, a human preference dataset containing 798k carefully annotated comparison pairs, enabling the training of Human Preference Score v2, the state-of-art preference prediction model, to better align text-to-image generation models' evaluation with human values and judgments.
We also provide a more stable and fair group of prompts for text-to-image generative models' evaluation, and set up a benchmark that compares a wide range of recent models. 
Finally, we illustrate the accuracy of HPS v2 by demonstrating some example usages. There are still many opportunities for further advancement and expansion based on our HPD v2 and HPS v2. 
We hope our work can inspire and facilitate future researches in this area.

\newpage

{
\small\bibliographystyle{ieee_fullname}
\bibliography{egbib.bib}
}

\section*{Checklist}

The checklist follows the references.  Please
read the checklist guidelines carefully for information on how to answer these
questions.  For each question, change the default \answerTODO{} to \answerYes{},
\answerNo{}, or \answerNA{}.  You are strongly encouraged to include a {\bf
justification to your answer}, either by referencing the appropriate section of
your paper or providing a brief inline description.  For example:
\begin{itemize}
  \item Did you include the license to the code and datasets? \answerYes{Both the code and dataset is released under Apache 2.0 License.}
  \item Did you include the license to the code and datasets? \answerNo{The code and the data are proprietary.}
  \item Did you include the license to the code and datasets? \answerNA{}
\end{itemize}
Please do not modify the questions and only use the provided macros for your
answers.  Note that the Checklist section does not count towards the page
limit.  In your paper, please delete this instructions block and only keep the
Checklist section heading above along with the questions/answers below.

\begin{enumerate}

\item For all authors...
\begin{enumerate}
  \item Do the main claims made in the abstract and introduction accurately reflect the paper's contributions and scope?
    \answerYes{}
  \item Did you describe the limitations of your work?
    \answerYes{}
  \item Did you discuss any potential negative societal impacts of your work?
    \answerNA{}
  \item Have you read the ethics review guidelines and ensured that your paper conforms to them?
    \answerYes{}
\end{enumerate}

\item If you are including theoretical results...
\begin{enumerate}
  \item Did you state the full set of assumptions of all theoretical results?
    \answerNA{}
	\item Did you include complete proofs of all theoretical results?
    \answerNA{}
\end{enumerate}

\item If you ran experiments (e.g. for benchmarks)...
\begin{enumerate}
  \item Did you include the code, data, and instructions needed to reproduce the main experimental results (either in the supplemental material or as a URL)?
    \answerYes{Please see the supplemental material.}
  \item Did you specify all the training details (e.g., data splits, hyperparameters, how they were chosen)?
    \answerYes{see Sec.~\ref{sec:Train}}
	\item Did you report error bars (e.g., with respect to the random seed after running experiments multiple times)?
    \answerYes{In Tab.~\ref{tab:benchmark}, we report the mean and standard deviation of 10 runs to demonstrate the stability and reliability of our benchmark.}
	\item Did you include the total amount of compute and the type of resources used (e.g., type of GPUs, internal cluster, or cloud provider)?
    \answerYes{The resource consumption will be reported in our GitHub repository.}
\end{enumerate}

\item If you are using existing assets (e.g., code, data, models) or curating/releasing new assets...
\begin{enumerate}
  \item If your work uses existing assets, did you cite the creators?
    \answerYes{}
  \item Did you mention the license of the assets?
    \answerNA{}
  \item Did you include any new assets either in the supplemental material or as a URL?
    \answerYes{We will release our dataset and pre-train model.}
  \item Did you discuss whether and how consent was obtained from people whose data you're using/curating?
    \answerYes{Please see Sec.~\ref{sec:preference_annotation}.}
  \item Did you discuss whether the data you are using/curating contains personally identifiable information or offensive content?
    \answerNA{}
\end{enumerate}

\item If you used crowdsourcing or conducted research with human subjects...
\begin{enumerate}
  \item Did you include the full text of instructions given to participants and screenshots, if applicable?
    \answerYes{We will show our instructions given to the workers in the supplemental material.}
  \item Did you describe any potential participant risks, with links to Institutional Review Board (IRB) approvals, if applicable?
    \answerNA{}
  \item Did you include the estimated hourly wage paid to participants and the total amount spent on participant compensation?
    \answerYes{Yes, the hourly wage for the works is 20CNY/hour for annotators and 25CNY/hour for checkers.}
\end{enumerate}

\end{enumerate}


\newpage
\appendix
\section{Prompt Cleaning}
\label{sec:appendix_cleaning}
\begin{figure}
    \centering
    \caption{Category Distribution.}
    \includegraphics[width=0.5\linewidth]{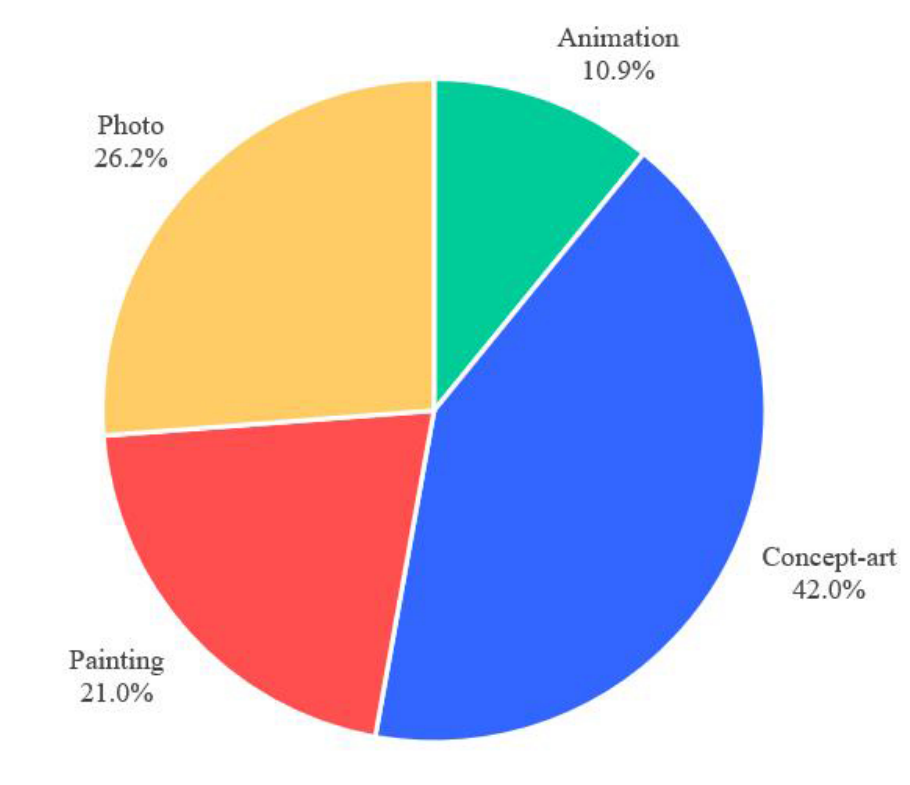}
    \label{fig:distribution}
\end{figure}

In this section we show details in prompt cleaning, categorization and annotation. As illustrated in Sec.~\ref{sec:prompt_collection}, we task ChatGPT to clean and categorize prompts by posing the following questions.
\begin{quote}
    - I will give you a description about an image.Remove modifiers from text that have nothing to do with the main content of the image, for example resolution, sharpness, light, image quality, authors and online platform, and describe it succinctly in one sentence.
\end{quote} 
\begin{quote}
    - Next, I will use text to describe a picture. Please reply to me according to the following rules:\\
    1. If the picture belongs to the style of ``paintings'', reply only with ``paintings'';\\
    2. If the picture belongs to the style of ``anime and cartoon'', reply only with ``anime and cartoon'';\\
    3. If the picture belongs to the style of ``real photo'', reply only with ``real photo'';\\
    4. If the picture belongs to the style of ``concept-art'', reply only with ``concept-art'';\\
    5. If the picture doesn't belong to any styles of above, reply only with ``others''; You must reply with only on word.
\end{quote}
Even though prompts of ``Photo'' category in HPD v2 are from COCO Captions~\cite{cococaptions}, we retain ``Photo'' in the classification process to mitigate the potential mistakes made by ChatGPT. The category distribution of HPD v2 is illustrated in Fig.~\ref{fig:distribution}.

Additionally, when processing data from DiffusionDB~\cite{wang2022diffusiondb}, we apply a filter to remove prompts with a NSFW threshold of 0.4. During the subsequent image generation step, we refine the selection of prompts by subjecting them to a stricter safety checker of DALL·E 2~\cite{dalle2}. This helps to further eliminate inappropriate vocabularies.

\begin{figure}[htbp]
    \centering
    \subfigure[]{
        \begin{minipage}[t]{0.5\linewidth}
        \centering
        \includegraphics[width=2in]{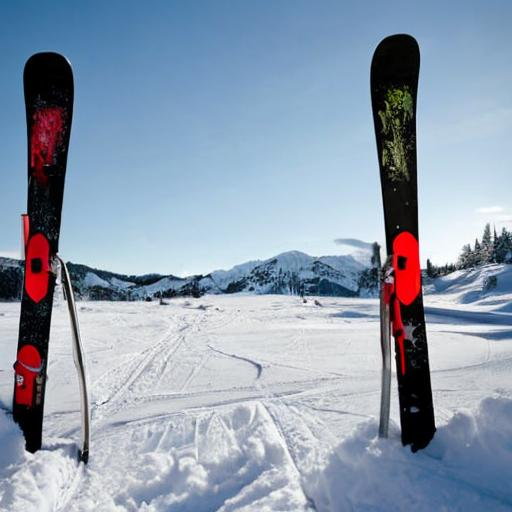}
        \label{exp1-a}
        \end{minipage}%
        }%
    \subfigure[]{
        \begin{minipage}[t]{0.5\linewidth}
        \centering
        \includegraphics[width=2in]{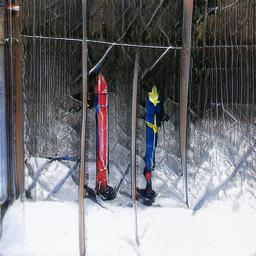}
        \label{exp1-b}
        \end{minipage}%
        }%
\centering
\caption{Prompt: A pair of skis standing up against a gate.}
\label{exp1}
\end{figure}

\section{Annotation}
\label{sec:appendix_annotation}
To ensure the quality of annotation, we provide a basic guidance document for our annotators:
\noindent \centerline{\textbf{Basic rules and regulations}} \\
We will provide you groups of images for ranking, which consists of images generated from different AI models. Please consider the prompts and rank images from the perspectives of universal and personal aesthetic appeal. This task mainly involves two aspects: text-image alignment and image quality. Although we encourage and value personal preference, it's important to consider the following fundamental principles when balancing the two aspects or facing a dilemma:
\begin{enumerate}
    \item
    When Image (A) surpasses Image (B) in terms of aesthetic appeal and fidelity, or Image (B) suffers from severe distortion and blurriness, even if Image (B) aligns better with the prompt, Image (A) should take precedence over Image (B). For example, in Fig.~\ref{exp1}, Fig.~\ref{exp1-b} lacks clear outlines and details of the ski board, resulting in an unattractive appearance and significant blurriness. However, Fig.~\ref{exp1-b} exhibits excellent fidelity and quality. Therefore, the ordering of the figures should place Fig.~\ref{exp1-a} before Fig.~\ref{exp1-b}.

\begin{figure}[htbp]
    \centering
    \subfigure[]{
        \begin{minipage}[t]{0.5\linewidth}
        \centering
        \includegraphics[width=2in]{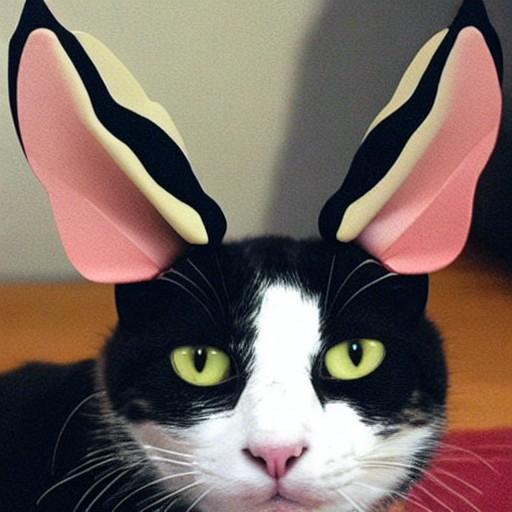}
        \label{exp2-a}
        \end{minipage}%
        }%
    \subfigure[]{
        \begin{minipage}[t]{0.5\linewidth}
        \centering
        \includegraphics[width=2in]{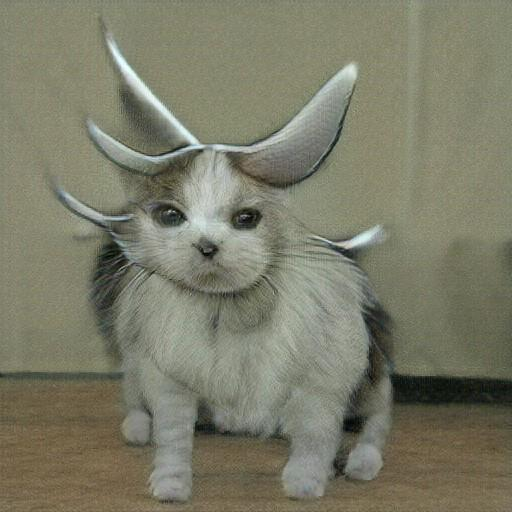}
        \label{exp2-b}
        \end{minipage}%
        }%
\centering
\caption{Prompt: A cat with two horns on its head.}
\label{exp2}
\end{figure}

    Similarly, in Fig.~\ref{exp2}, although in Fig.~\ref{exp2-a} the required horns are mistakenly generated as elf ears, Fig.~\ref{exp2-b} suffers severe structural issues, repetitive generation problems, and blurriness. Therefore, the ordering of the figures should place Fig.~\ref{exp2-a} before Fig.~\ref{exp2-b}.

    \item 
    When facing a dilemma that images are relatively similar in terms of aesthetics and personal preference, please carefully read and consider the prompt for sorting based more on the text-image alignment. For example, if you cannot make a choice based on personal preference, as in Fig.~\ref{exp3}, please pay attention to the description, which refers to a mouse mechanic. Therefore, the ordering of the figures should place Fig.~\ref{exp3-b} before Fig.~\ref{exp3-a}. 

\begin{figure}[htbp]
    \centering
    \subfigure[]{
        \begin{minipage}[t]{0.5\linewidth}
        \centering
        \includegraphics[width=2in]{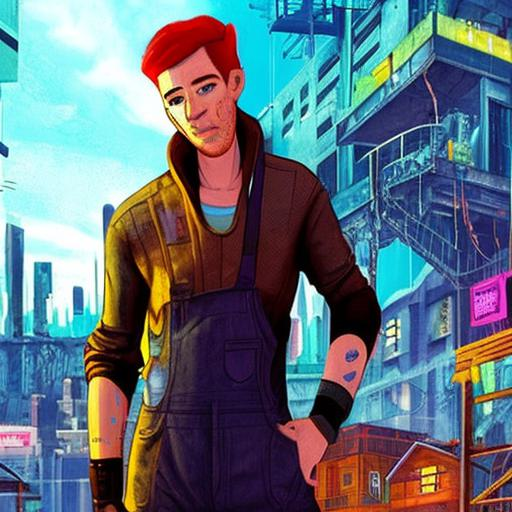}
        \label{exp3-a}
        \end{minipage}%
        }%
    \subfigure[]{
        \begin{minipage}[t]{0.5\linewidth}
        \centering
        \includegraphics[width=2in]{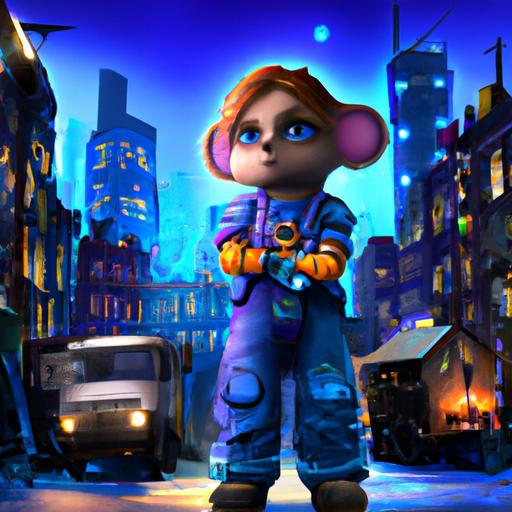}
        \label{exp3-b}
        \end{minipage}%
        }%
\centering
\caption{Prompt: A ginger haired mouse mechanic in blue overalls in a cyberpunk scene with neon slums in the background.}
\label{exp3}
\end{figure}

    \item 
    It is crucial to pay special attention to the capitalized names, as these names may lead to misunderstandings during the machine translation process. If there is any incorrectly translated proprietary term or content you are not familiar with, we recommend you to search for sample images and explanations online.

\end{enumerate}

\section{CLIP Details}
\label{sec:clip_details}
CLIP~\cite{clip} is a vision-language model trained on massive, noisy image-text pairs from the internet to model the joint distribution between images and texts.
They demonstrate that the simple pre-training task
of predicting which caption corresponds to which image is an efficient and scalable way to learn
image representations from scratch~\cite{clip}.
The learned feature demonstrates impressive zero-shot image classification accuracy on various tasks.
Our task of predicting human preference can also be viewed as an image classification task, so it can naturally take advantage of the CLIP pre-training.
We use the ViT-H/14 variant of CLIP pre-trained by the OpenClip~\cite{openclip} project on LAION-2B~\cite{schuhmann2022laionb}.
The model consists of an image transformer and a text transformer as encoders for both modality.
Details about the model architecture is listed in Tab.~\ref{tab:model_card}.
\input{tables/clip_model_card}

\section{Generation Details}
\label{sec:generation_details}
We provide image generation details for each model when constructing the HPS v2 benchmark.
The images are also released with HPD v2.

\noindent \textbf{Stable Diffusion.} For Stable Diffusion v1.4 and v2.0, we generate images of size $512\times512$ using a guidance scale of 7.5 for 50 steps with DDIM sampler.

\noindent \textbf{GLIDE.} Images of size $256\times256$ are generated from the official mini-glide model with a guidance scale of 3.0 for 27 steps with DDIM sampler.

\noindent \textbf{LAFITE.} Images of size $256\times256$ are generated following default configuration using model\footnote{https://drive.google.com/file/d/1tMD6MWydRDMaaM7iTOKsUK-Wv2YNDRRt/view?usp=sharing}.

\noindent \textbf{VQ-Diffusion.} Images of size $256\times256$ are generated with a guidance scale of 5.0 and 100 sampling steps with VQ-Diffusion sampler using model\footnote{https://huggingface.co/microsoft/vq-diffusion-ithq}.

\noindent \textbf{FuseDream.} Images of size $512\times512$ are generated with 1000 initialization iterations, 1000 optimization iterations and basis image number is set to 5.

\noindent \textbf{Latent Diffusion.} Images are generated from the official LDM model with the DPM Solver sampler, 25 sampling steps, $\eta=0.3$, and a guidance scale of 6.0.

\noindent \textbf{CogView2.} Images of size $512\times512$ are generated with default hyper-parameters. For category ``animation'', ``concept-art'', ``painting'' and ``photo'' in HPD v2, we set the ``style'' in Cogview2 to ``none'', ``none'', ``oil'' and ``photo'' respectively. 

\noindent \textbf{DALL·E 2.} Images are directly collected by requesting the official API\footnote{https://openai.com/}.

\noindent \textbf{Versatile Diffusion.} Images are generated from the official model using the DPM Solver sampler, 25 sampling steps, and a guidance scale of 7.5.

\noindent \textbf{VQGAN+CLIP.} Images of size $512\times512$ are generated with 500 iterations and Adam optimizer with a learning rate of 0.1.

\noindent \textbf{Community Models based on Stable Diffusion (Epic Diffusion, Openjourney, MajicMix Realistic, ChilloutMix, Deliberate, Realistic Vision, Dreamlike Photoreal 2.0)} Images of size $512\times512$ are generated from the publicly released model with the DPM Solver sampler, 25 sampling steps, and a guidance scale of 7.5.

\noindent \textbf{DeepFloyd-XL} Stages 1 and 2 are set to use DDPM noise schedulers with 25 sampling steps and the default samplers and guidance scales. Then the generated $256\times256$ images are resized to $192\times192$. For the upsampling stage, we use Stable Diffusion upscaler, with the DPM Solver sampler and 25 steps to generate $768\times768$ images.

\noindent \textbf{DALL·E mini} We use the fast and minimal reproduction of the model\footnote{https://huggingface.co/kuprel/min-dalle} and all hyper-parameters are set to default except $grid\_size=1$.

\input{tables/pairwise}

\section{Pairwise Accuracy}
\label{sec:pairwise_acc}
\begin{figure}[ht]
    \caption{Chrominance and luminance distribution of images in HPD v2 test split}
    \label{fig:chrominance}
    \begin{minipage}[b]{1.0\textwidth}
        \centering
        \includegraphics[width=1.0\linewidth]{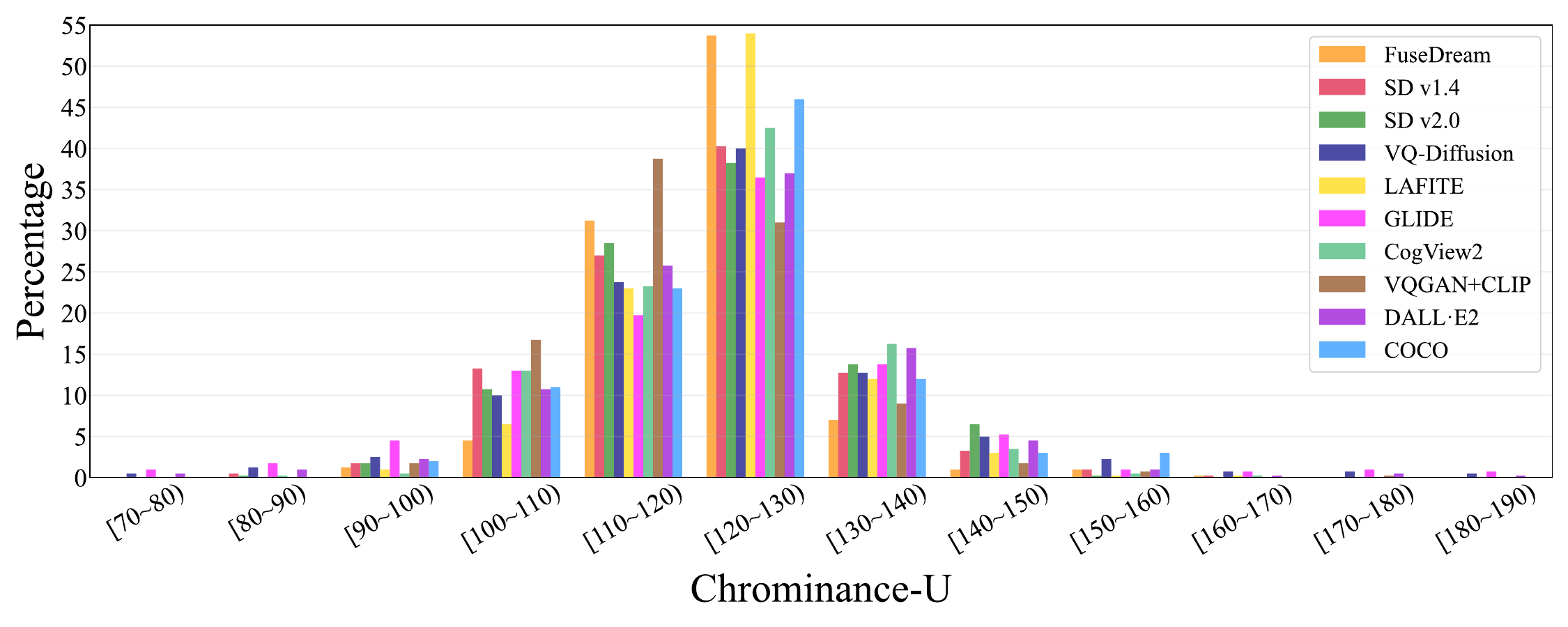}
        \label{fig:chrominance_u}
    \end{minipage}%
    \\
    \begin{minipage}[b]{1.0\textwidth}
        \centering
        \includegraphics[width=1.0\linewidth]{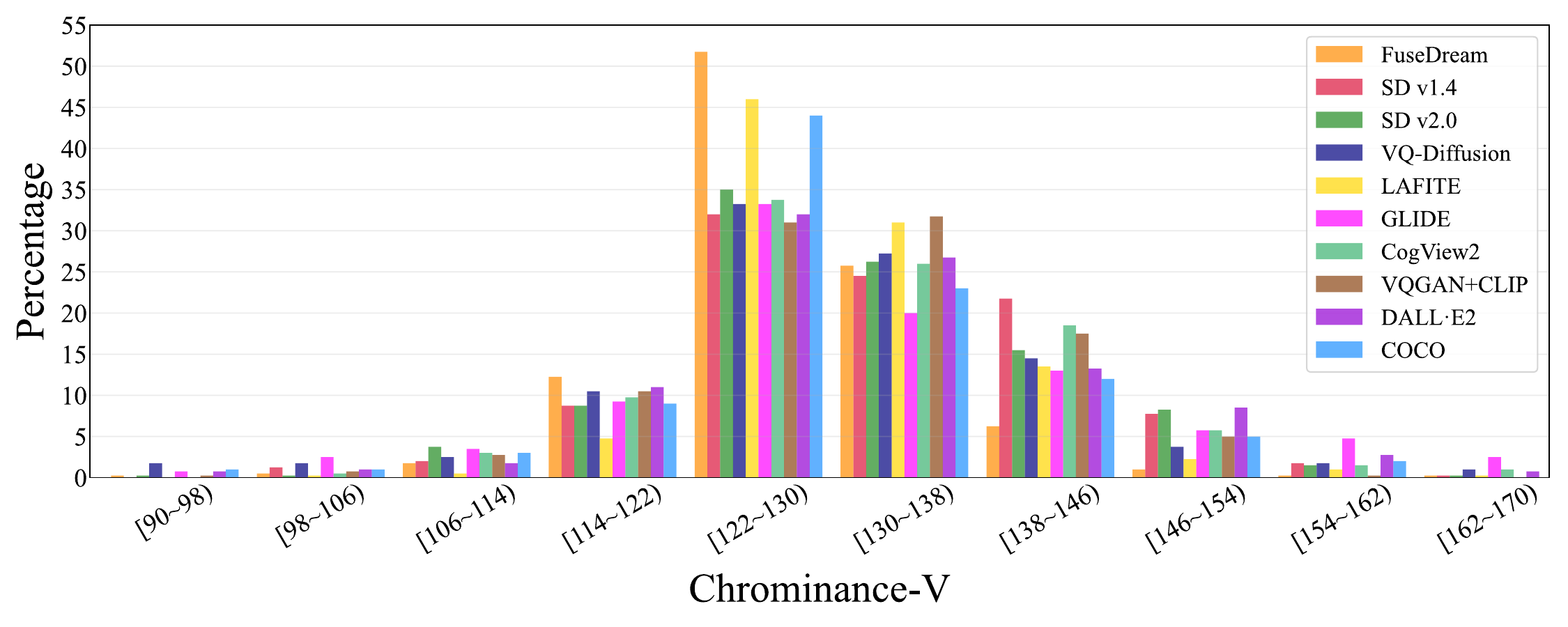}
        \label{fig:chrominance_v}
    \end{minipage}
    \begin{minipage}[b]{1.0\textwidth}
        \centering
        \label{fig:luminance}
        \includegraphics[width=1.0\linewidth]{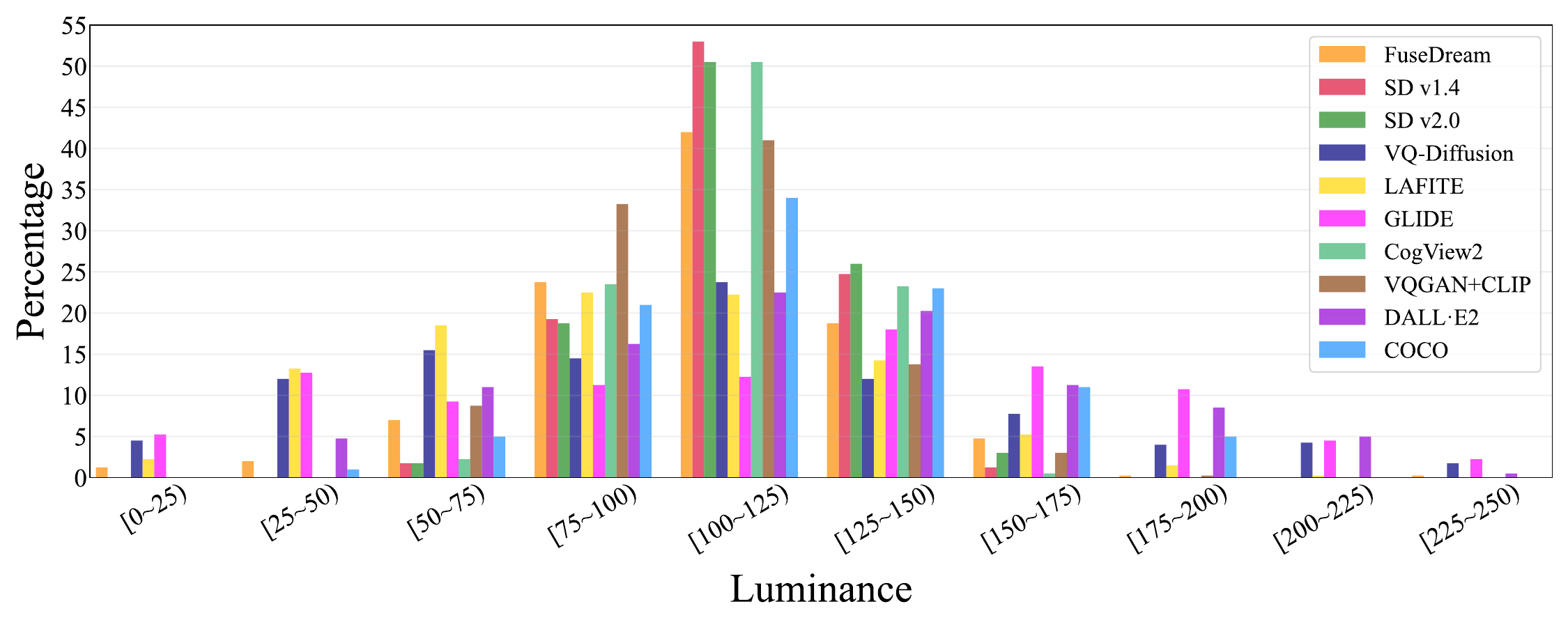}
    \end{minipage}
\end{figure}

To further illustrate the accuracy of HPS v2, we evaluate its agreement with human choices between all pairs of generative models in test split. As shown in Tab.~\ref{tab:pairwise eval}, the predictions of our model align closely with human choices between most models.
We also show the chrominance and luminance distribution of each model in HPD v2 test split in Fig.~\ref{fig:chrominance} as a reference.

\section{Ablation Study}
\label{sec:ablations}

We ablate hyper-parameter choices of training HPS v2 on Tab.~\ref{tab:ablation_lr}, Tab.~\ref{tab:ablation_image}, Tab.~\ref{tab:ablation_text}.

\begin{table}[ht]
    \begin{minipage}{.28\linewidth}
        \vspace{0.5cm}
        \caption{Ablation studies on learning rates.}
        \label{tab:ablation_lr}
        \centering
        \resizebox{1.0\linewidth}{!}{
            \input{tables/ablation_lr}
        }
    \end{minipage}
    \hfill
    \begin{minipage}{.3\linewidth}
        \caption{Ablation studies on \# of free layers in the image encoder.}
        \label{tab:ablation_image}
        \centering
        \resizebox{1.0\linewidth}{!}{
            \input{tables/ablation_image}
        }
    \end{minipage}
    \hfill
    \begin{minipage}{.29\linewidth}
        \caption{Ablation studies on \# of free layers in the text encoder.}
        \label{tab:ablation_text}
        \centering
        \resizebox{1.0\linewidth}{!}{
            \input{tables/ablation_text}
        }
    \end{minipage}
\end{table}

\section{Example Usages of HPS v2}
\label{sec:example_usages}
We show 2 example usages of HPS v2 in this section to show its sensitivity and accuracy.

\begin{figure}
    \centering
    \includegraphics[width=0.8\linewidth]{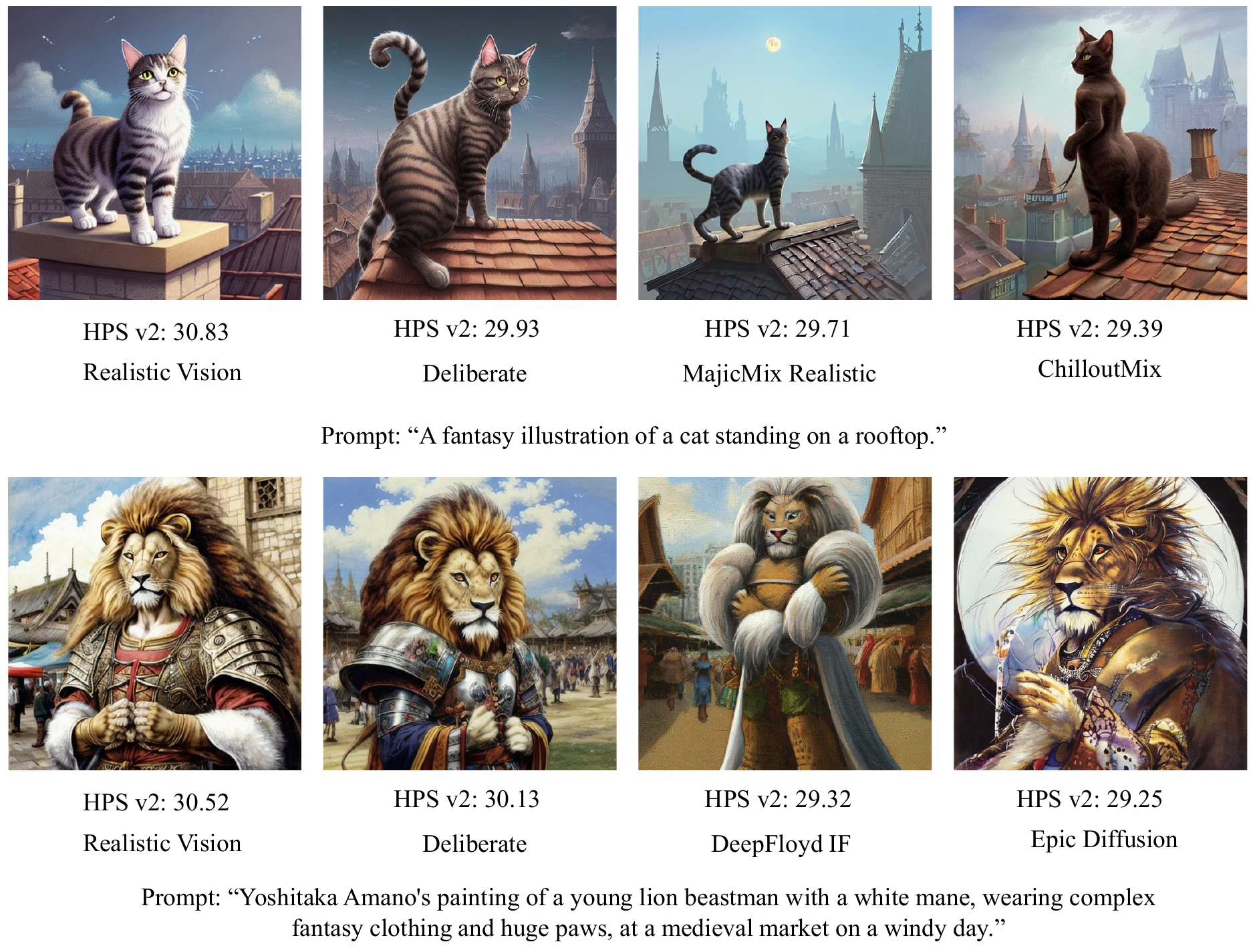}
    \caption{Examples of HPS v2 evaluated on community models.}
    \label{fig:enter-label}
\end{figure}

\begin{figure}
    \centering
    \includegraphics[width=1.0\linewidth]{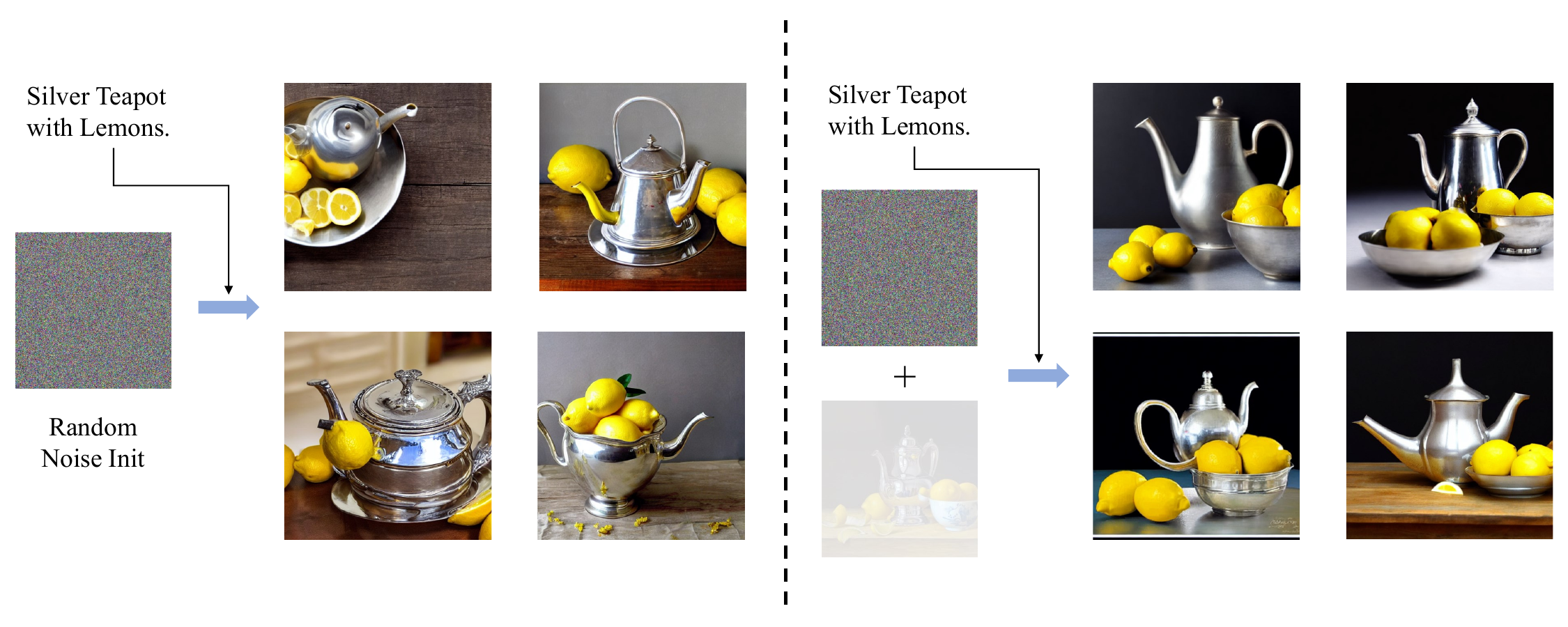}
    \caption{Left: the standard denoising inference. Right: blending the random initial noise with a prior image. The input random noise has strong influence on the layout of the generated image.}
    \label{fig:noise}
\end{figure}

\subsection{Retrieval initialization}
\noindent\textbf{Method.}
Many recent works~\cite{samuel2023all} suggest that the latent feature initialization has a great influence on the output image quality of Stable Diffusion.
We observe that this may relate to a misalignment between the noise schedule of training and inference process.
We find that when training Stable Diffusion, there isn't a timestep at which the input latent feature is fully random, but at inference time the image is recovered from random noise, which is inconsistent with the training stage.
This implies that the inference of Stable Diffusion can be potentially improved by initializing with better latent features.
With this observation, we try a straightforward idea of blending the latent feature of a reference image with random noise to initialize the latent features.

Given a prompt, a prior image is generated by a powerful community model Dreamlike Photoreal 2.0, and then the image is encoded into latent space with noise level similar to the training stage.
The noisy latent feature is used to initialize the inference of Stable Diffusion, which is previously sampled from $\mathcal{N}(\mathbf{0}, \mathbf{I})$ without prior.

\textbf{Results. }
We verify the effectiveness of this method by comparing it with the default initialization of Stable Diffusion. 
Tab.\ref{tab:example_usages} demonstrates that the improvement brought by initializing via retrieved images can be verified by HPS v2.
Visualizations in Fig.\ref{fig:noise} also show that the noise initialization has a strong impact on the global layout of the generated image, thus randomly initializing it may lead to undesired image layouts. 

\subsection{Evaluating adapted model in HPS v1}
Wu et al.\cite{wu2023better} improves Stable Diffusion to better align it with human preference, as introduced in HPS v1.
However, the improvement is only validated through user studies. 
There are two reasons that prevent the authors from evaluating via HPS v1.
Firstly, the bias of HPS v1 and the evaluation prompts are under-studied, so HPS v1 is not ready for serving as an evaluation metric. 
Secondly, HPS v1 also incorporates the training of the adapted model. 
With HPS v2, we are now able to quantify the actual improvement of the method, as shown in Tab.\ref{tab:example_usages}.

\input{tables/example_usages}

\section{Datasheet}

\subsection{\centering \centering Motivation}
\subsection*{Why was the dataset created?} 
\noindent The dataset was created to facilitate future academic Computer Vision research about human aesthetic preference.
\subsection*{Who created this dataset (\eg which team, research group) and on behalf of which entity (\eg company, institution, organization)?}
\noindent The dataset was created by researchers at MMLab, The Chinese University of Hong Kong and Centre for Perceptual and Interactive Intelligence (CPII).

\subsection{\centering Composition}
\subsection*{What do the instances that comprise the dataset represent (\eg
documents, photos, people, countries)? Are there multiple types of instances? (\eg movies, users, ratings; people, interactions between them; nodes, edges)}
\noindent
The instances are prompts and generated images, along with human preference choices among the images generated by the same prompt.

\subsection*{Are relationships between instances made explicit in the data (\eg social network links, user/movie ratings, etc.)?}
\noindent
No, we did not track the annotator identity of each instance.

\subsection*{How many instances are there? (of each type, if appropriate)?}
\noindent 
There are 789,090 instances in the dataset, standing for 789,090 comparisons between image pairs. Specifically, our dataset has 433,760 images and 107,515 prompts. 
The benchmark contains 3200 unique prompts without paired images.

\subsection*{What data does each instance consist of? ``Raw'' data (\eg unprocessed text or images) or Features/attributes? Is there a label/target associated with instances? If the instances related to people, are sub-populations identified (\eg by age, gender, etc.) and what is their distribution?}
\noindent 
Each instance consists of two images, one prompt and one human choice. 

\subsection*{Is any information missing from individual instances? If so, please
provide a description, explaining why this information is missing (\eg
because it was unavailable). This does not include intentionally removed
information, but might include, \eg redacted text.}
\noindent 
Yes, we omit the specific parameters for generating the images, such as diffusion steps and guidance scale.
They are omitted because we are more interested in the users' preference about the generated images, rather than how they are created. Instead, we illustrate the hyper-parameters used for generation in Appendix~\ref{sec:generation_details}.

\subsection*{Is everything included or does the data rely on external resources?}
\noindent The dataset is self-contained.

\subsection*{Are there recommended data splits and evaluation measures? (\eg training, development, testing; accuracy or AUC)}
\noindent 
For the preference model training, we use a training set of 645,090 instances and a validation set of 153,000 instances.
We recommend using accuracy (\%) with 1 decimal place.
For evaluating text-to-image generative models, we recommend reporting HPS v2 with 4 decimal place, as well as the standard deviation.

\subsection*{Are there any errors, sources of noise, or redundancies in the
dataset?}
\noindent There is potential noise in the dataset. Since human preference is highly diverse, in most cases, it's hard to tell whether the divergence in human choice is due to noise or disagreement of aesthetic criterion.

\subsection*{Is the dataset self-contained, or does it link to or otherwise rely on external resources (\eg websites, tweets, other datasets)?}
\noindent The dataset is self-contained.

\subsection*{Does the dataset contain data that might be considered confidential (\eg data that is protected by legal privilege or by doctorpatient confidentiality, data that includes the content of individuals non-public communications)?}
\noindent 
No, both prompt sources (DiffusionDB and COCO Captions) and image generative models are publicly available. 

\subsection*{Does the dataset contain data that, if viewed directly, might be offensive, insulting, threatening, or might otherwise cause anxiety? If so, please describe why.}
\noindent
We retained the prompts with NSFW scores below a threshold and accepted by DALL·E2. However, there may still be an extremely small portion of inappropriate content that were not removed by our policy.

\subsection*{Does the dataset relate to people?}
\noindent Yes, the choices are made by human annotators.

\subsection*{Does the dataset identify any subpopulations (\eg by age, gender)?}

\noindent No, but we report the statistics of subpopulations by age, gender and occupancy.

\subsection*{Is it possible to identify individuals (\ie one or more natural persons), either directly or indirectly (\ie in combination with other data) from the dataset?}
\noindent No.

\subsection*{Does the dataset contain data that might be considered sensitive in any way (\eg data that reveals racial or ethnic origins, sexual orientations, religious beliefs, political opinions or union memberships, or locations; financial or health data; biometric or genetic data; forms of government identification, such as social security numbers; criminal history)?}

The dataset may contain sensitive data, because the prompts are sourced from DiffusionDB, which are collected from users. Thus, it may contain sensitive information, such as public figures and religious beliefs.

\subsection*{What experiments were initially run on this dataset? Have a summary of those results.}
\noindent 
We firstly collected the test split to evaluate existing preference prediction models and found that they do not align well with human choices. Details are illustrated in Tab.~\ref{tab:evaluation}. Based on this, we collected the large scale training split of HPD v2 and train a more aligned preference prediction model.


\subsection{\centering Data Collection Process}

\subsection*{How was the data associated with each instance acquired?}

\noindent
Each instance contain two images generated from the same prompt, and a human choice annotation.

\subsection*{What mechanisms or procedures were used to collect the data (\eg hardware apparatus or sensor, manual human curating, software program, software API)?}
\noindent We use ChatGPT API provided by OpenAI to process prompts from DiffusionDB and an annotation system.

\subsection*{If the dataset is a sample from a larger set, what was the sampling strategy (\eg deterministic, probabilistic with specific sampling probabilities)?}

\noindent  The dataset is not a sample of a larger set. 

\subsection*{Who was involved in the data collection process (\eg students, crowd-workers, contractors) and how were they compensated (\eg
how much were crowdworkers paid)?}

\noindent  We hire 57 annotators for human preference annotation, but they prefer not to disclose their salary information due to personal privacy reasons.

\subsection*{Over what time-frame was the data collected?}
\noindent 
The dataset was collected between 2023-03-15 and 2023-06-03.

\subsection*{Were any ethical review processes conducted (\eg by an institutional review board)?}
\noindent No official process is conducted.

\subsection*{Does the dataset relate to people?}
\noindent No. Although crowdsourcing was used for annotation, we didn't perform any kind of intervention or interaction during data collection and didn't collect any personal information or biospecimens of annotators. It's not possible to identify individuals from our dataset. In addition, according to the definition of ``Human Subjects Research'' given by NIH(National Institutes of Health of USA)\footnote{https://grants.nih.gov/policy/humansubjects/research.htm}, our research is not regarded as ``research with human subjects''.

\subsection*{Did you collect the data from the individuals in question directly, or obtain it via third parties or other sources (\eg websites)?}
\noindent 
We collected human choices directly from annotators in question.





\subsection*{Has an analysis of the potential impact of the dataset and its use on data subjects (\eg a data protection impact analysis) been conducted?}

\noindent No analysis has been conducted.


\subsection{\centering Data Preprocessing}
\subsection*{What preprocessing/cleaning was done? (\eg discretization or bucketing, tokenization, part-of-speech tagging, SIFT feature extraction, removal of instances, processing of missing values)?}
\noindent We conducted prompt cleaning process using ChatGPT to remove biased functional words. Please see Sec.~\ref{sec:appendix_cleaning} and Sec.~\ref{sec:prompt_collection} for details.




\subsection{\centering Uses}
\subsection*{Has the dataset been used for any tasks already? If so, please provide a description.}
\noindent As described in the paper, this dataset has been used for training HPS v2.

\subsection*{Is there a repository that links to any or all papers or systems that use the dataset?}
\noindent
No.

\subsection*{What (other) tasks could the dataset be used for?}
\noindent 
It can be used for tasks related to human preference on generated images.

\subsection*{Is there anything about the composition of the dataset or the way it was collected and preprocessed/cleaned/labeled that might impact future uses?}
\noindent Yes. 
As discussed in Sec.~\ref{sec:limitations}, there might be bias induced by the limited number of annotators.


\subsection*{Are there tasks for which the dataset should not be used?}
\noindent No.


\subsection{\centering Data Distribution}
\subsection*{Will the dataset be distributed to third parties outside of the entity (\eg company, institution, organization) on behalf of which
the dataset was created? If so, please provide a description}
\noindent Yes. Anyone can access the dataset via the internet.

\subsection*{How will the dataset be distributed? (\eg tarball on website, API, GitHub; does the data have a DOI and is it archived redundantly?)}
\noindent We will provide a download link in a GitHub repository. 

\subsection*{When will the dataset be distributed?}
\noindent When the paper is accepted.

\subsection*{Will the dataset be distributed under a copyright or other intellectual property (IP) license, and/or under applicable terms of use (ToU)?}
\noindent We will provide a terms-of-use agreement with the dataset. The dataset as a whole is distributed under Apache 2.0 license.

\subsection*{Have any third parties imposed IP-based or other restrictions on
the data associated with the instances? If so, please describe these restrictions, and provide a link or other access point to, or otherwise reproduce, any relevant licensing terms, as well as any fees associated
with these restrictions.}
\noindent No.

\subsection*{Do any export controls or other regulatory restrictions apply to
the dataset or to individual instances? If so, please describe these
restrictions, and provide a link or other access point to, or otherwise
reproduce, any supporting documentation.}
\noindent Unknown.


\subsection{\centering Dataset Maintenance}
\subsection*{Who is supporting/hosting/maintaining the dataset?}
\noindent The authors of this paper are maintainers of this dataset.

\subsection*{How can the owner/curator/manager of the dataset be contacted
(\eg email address)?}
\noindent By email: wuxiaoshi@link.cuhk.edu.hk.

\subsection*{Is there an erratum?}
\noindent At this time, we are not aware of errors in our dataset. However, we will create an erratum as errors are identified.

\subsection*{Will the dataset be updated? If so, how often and by whom? 
How will updates be communicated? (\eg mailing list, GitHub)}
\noindent 
Yes, the dataset will be updated when necessary.
The dataset will be updated when:
\begin{itemize}
    \item inappropriate content is spotted in the dataset.
    \item HPS v2 loses consistency with human preferences for new generative models.
    \item other improvements are available.
\end{itemize}
Updates will be recorded in the GitHub repository\footnote{https://github.com/tgxs002/HPSv2}.


\subsection*{Will older versions of the dataset continue to be supported/hosted/maintained?}
\noindent Yes.

\subsection*{If others want to extend/augment/build on this dataset, is there a mechanism for them to do so? If so, is there a process for tracking/assessing the quality of those contributions. What is the process for communicating/distributing these contributions to users?}
\noindent The authors reserve the right to maintain the dataset.

\end{document}

%% file: tables/prompt_cleaning.tex
\begin{table}
  \caption{Examples of prompts cleaned by ChatGPT. }
  \label{tab:clean_prompt}
  \centering
  \resizebox{0.8\linewidth}{!}{
\begin{tabularx}{5in}{XX}
\toprule
Prompts from DiffusionDB~\cite{wang2022diffusiondb} &  Prompts cleaned by ChatGPT \\
\midrule
\RaggedRight{queen goddess isis, high fantasy, by john berkey, by peter mohrbacher, renaissance oil painting, realism, master study, hyperrealist.}&
\RaggedRight{The image depicts a regal depiction of the Egyptian goddess Isis.}\\
\midrule
\RaggedRight{
star wars portrait of a robert redford by greg rutkowski, jacen solo, very sad and relucant expression, wearing a biomechanical suit, scifi, digital painting, artstation, concept art, smooth, artstation hq. 
} & 
\RaggedRight{
A digital painting of Star Wars character Jacen Solo in a biomechanical suit with a sad expression.
} \\
\bottomrule
\end{tabularx}
}
\end{table}

%% file: tables/source_model.tex
  \begin{tabular}{lllll}
    \toprule
     Source & Split & \# Param. & Type & \# images \\ 
     \midrule
     CogView2 & train \& test & 24B & Autoregressive & 73697 \\ 
     DALL·E 2 & train \& test & 3.5B & Diffusion & 101869 \\ 
     GLIDE & test & 0.94B & Diffusion & 400 \\ 
     SD v1.4 & train \& test & 0.89B & Diffusion  & 101869\\ 
     SD v2.0 & train \& test & 0.89B & Diffusion  & 101869 \\ 
     LAFITE & test & 0.75B & GAN & 400 \\ 
     VQGAN+CLIP & test & 0.73B & GAN & 400  \\ 
     VQ-Diffusion & test & 0.37B & Diffusion & 400 \\
     FuseDream & test & 0.35B & GAN & 400 \\ 
     COCO Captions & train \& test & - & - & 28272\\
    \bottomrule
  \end{tabular}

%% file: tables/annotator.tex
\begin{tabular}{llc}
    \toprule
    \multirow{2}{*}{Gender}& Male & 25 \\ & Female & 32 \\
    \hline
    \multirow{3}{*}{Education}& University & 8 \\ & College & 42 \\& Before college & 7 \\
    \hline
    \multirow{3}{*}{Age}& 18-25 & 13 \\ & 25-35 & 37 \\ & 35 or above & 7 \\
    \hline
    \multirow{2}{*}{Employment}& Full time & 35 \\ & Part time & 22 \\
    \midrule
    Total & - & 57 \\
    \bottomrule
\end{tabular}

%% file: tables/human_vs_human.tex
\begin{table}
  \caption{Inter-annotator agreement of HPD v2 annotation between different models.}
  \label{tab:huamn_vs_human}
  \centering
  \resizebox{1.0\textwidth}{!}{
  \begin{tabular}{lccccccccc}
    \toprule
     Model & SD v1.4 & SD v2.0 & VQ-Diffusion & LAFITE & GLIDE & CogView2 & VQGAN+CLIP & DALL·E2 & COCO \\ 
     \midrule
     FuseDream & 79.1\% & 83.6\% & 63.1\% & 69.1\% & 80.6\% & 78.1\% & 61.6\% & 82.2\% & 85.2\% \\
     SD v1.4 & - & 65.2\% & 82.4\% & 88.3\% & 92.2\% & 62.1\% & 81.9\% & 66.0\% & 71.0\% \\
     SD v2.0 & - & - & 86.9\% & 90.6\% & 93.1\% & 65.4\% & 85.4\% & 66.0\% & 65.8\% \\
     VQ-Diffusion & - & - & - & 72.8\% & 82.5\% & 77.9\% & 65.3\% & 80.8\% & 84.4\% \\
     LAFITE & - & - & - & - & 76.8\% & 85.0\% & 66.5\% & 86.4\% & 83.0\% \\
     GLIDE & - & - & - & - & - & 89.6\% & 77.2\% & 91.5\% & 90.2\% \\
     CogView2 & - & - & - & - & - & - & 80.0\% & 65.7\% & 68.1\% \\
     VQGAN+CLIP & - & - & - & - & - & - & - & 82.5\% & 88.2\% \\
     DALL·E 2 & - & - & - & - & - & - & - & - & 67.3\% \\
    \bottomrule
  \end{tabular}
}
\end{table}

%% file: tables/dataset_comparison.tex
\begin{table}
  \caption{Comparison with related datasets.}
  \label{tab:dataset_comparison}
  \centering
  \resizebox{1.0\textwidth}{!}{
  \begin{tabular}{lllllll}
    \toprule
     Name & Annotator & Data Format & Prompt Source & Image Source & \# Prompts & \# Choices \\ 
     \midrule
     HPD v1~\cite{wu2023better} & Discord users & Top-1 choice & Discord users & Stable Diffusion & 25k & 25k \\
     ImageReward~\cite{xu2023imagereward} & Expert & Pairwise comparison & DiffusionDB~\cite{wang2022diffusiondb} & DiffusionDB~\cite{wang2022diffusiondb} & 9k & 137k \\ 
     Pick-a-Pic~\cite{kirstain2023pick} & Web application users & Pairwise comparison & Web application users & 4 Models & 38k & 584k \\
     HPD v2 (ours) & Expert & Pairwise comparison & DiffusionDB~\cite{wang2022diffusiondb} cleaned by ChatGPT & 9 Models + real photo & 108k & 798k \\
    \bottomrule
  \end{tabular}
  }
\end{table}

%% file: tables/evaluation.tex
\begin{table}
  \caption{Preference prediction accuracy on test sets of ImageReward and HPD v2.}
  \label{tab:evaluation}
  \centering
  \resizebox{0.6\textwidth}{!}{
  \begin{tabular}{llccc}
    \toprule
    Model & ImageReward & HPD v2 \\
    \midrule
    CLIP ViT-H/14~\cite{clip, openclip} & 57.1 & 65.1 \\
    ImageReward~\cite{xu2023imagereward} & 65.1 & 74.0 \\
    Aesthetic Score Predictor~\cite{schuhmann2022laion} & 57.4 & 76.8 \\
    HPS~\cite{wu2023better} & 61.2 & 77.6 \\
    PickScore~\cite{kirstain2023pick} & 62.9 & 79.8 \\
    \midrule
    Single Human vs. Single Human & 65.3 & 78.1 \\
    Single Human vs. Averaged Human & 53.9 & 85.0 \\
    \midrule
    HPS v2 & \textbf{65.7} & \textbf{83.3} \\
    \bottomrule
  \end{tabular}
  }
\end{table}

%% file: tables/benchmark.tex
\begin{table}
  \caption{HPS v2 benchmark. We also provide HPS v2 evaluated on DrawBench for reference.}
  \label{tab:benchmark}
  \centering
  \resizebox{1.0\textwidth}{!}{
  \begin{tabular}{lccccc}
    \toprule
     & \multicolumn{2}{r}{HPS v2}                   \\
    \cmidrule(r){2-5}
    Model & Animation & Concept-art & Painting & Photo & DrawBench~\cite{imagen}\\
    \midrule
    GLIDE~\cite{glide} & $23.34\pm0.198$ & $23.08\pm0.174$ & $23.27\pm0.178$ & $24.50\pm0.290$ & $25.05\pm0.84$ \\
    LAFITE~\cite{zhou2021lafite} & $24.63\pm0.101$ & $24.38\pm0.087$ & $24.43\pm0.155$ & $25.81\pm0.213$ & $25.23\pm0.72$ \\
    VQ-Diffusion~\cite{gu2022vector} & $24.97\pm0.186$ & $24.70\pm0.149$ & $25.01\pm0.145$ & $25.71\pm0.222$ & $25.44\pm0.83$ \\
    FuseDream~\cite{liu2021fusedream} & $25.26\pm0.125$ & $25.15\pm0.107$ & $25.13\pm0.183$ & $25.57\pm0.248$ & $25.72\pm0.71$ \\
    Latent Diffusion~\cite{stable_diffusion} & $25.73\pm0.125$ & $25.15\pm0.140$ & $25.25\pm0.178$ & $26.97\pm0.183$ & $26.17\pm0.85$ \\
    CogView2 ~\cite{ding2022cogview2} & $26.50\pm0.129$ & $26.59\pm0.119$ & $26.33\pm0.100$ & $26.44\pm0.271$ & $26.17\pm0.74$ \\
    DALL·E mini & $26.10\pm0.132$ & $25.56\pm0.137$ & $25.56\pm0.112$ & $26.12\pm0.233$ & $26.34\pm0.76$ \\
    Versatile Diffusion~\cite{xu2022versatile} & $26.59\pm0.178$ & $26.28\pm0.145$ & $26.43\pm0.102$ & $27.05\pm0.229$ & $26.77\pm0.78$ \\
    VQGAN + CLIP~\cite{vqgan} & $26.44\pm0.152$ & $26.53\pm0.075$ & $26.47\pm0.111$ & $26.12\pm0.210$ & $26.38\pm0.43$ \\
    DALL·E 2~\cite{dalle2} & $27.34\pm0.175$ & $26.54\pm0.127$ & $26.68\pm0.156$ & $27.24\pm0.198$ & $27.16\pm0.64$ \\
    Stable Diffusion v1.4~\cite{stable_diffusion} & $27.26\pm0.156$ & $26.61\pm0.082$ & $26.66\pm0.143$ & $27.27\pm0.226$ & $27.23\pm0.57$ \\
    Stable Diffusion v2.0~\cite{stable_diffusion} & $27.48\pm0.174$ & $26.89\pm0.076$ & $26.86\pm0.120$ & $27.46\pm0.198$ & $27.31\pm0.68$ \\
    Epic Diffusion & $27.57\pm0.163$ & $26.96\pm0.113$ & $27.03\pm0.088$ & $27.49\pm0.192$ & $27.33\pm0.63$ \\
    Openjourney & $27.85\pm0.145$ & $27.18\pm0.090$ & $27.25\pm0.124$ & $27.53\pm0.178$ & $27.44\pm0.66$ \\
    MajicMix Realistic & $27.88\pm0.197$ & $27.19\pm0.094$ & $27.22\pm0.149$ & $27.64\pm0.176$ & $27.47\pm0.64$ \\ 
    ChilloutMix & $27.92\pm0.131$ & $27.29\pm0.090$ & $27.32\pm0.154$ & $27.61\pm0.195$ & $27.47\pm0.57$ \\
    DeepFloyd-XL & $27.64\pm0.108$ & $26.83\pm0.137$ & $26.86\pm0.131$ & $27.75\pm0.171$ & $27.64\pm0.72$ \\
    Deliberate & $28.13\pm0.135$ & $27.46\pm0.098$ & $27.45\pm0.111$ & $27.62\pm0.201$ & $27.73\pm0.64$ \\
    Realistic Vision & $28.22\pm0.133$ & $27.53\pm0.093$ & $27.56\pm0.124$ & $27.75\pm0.226$ & $27.77\pm0.64$ \\
    Dreamlike Photoreal 2.0 & $28.24\pm0.143$ & $27.60\pm0.085$ & $27.59\pm0.110$ & $27.99\pm0.245$ & $27.88\pm0.75$ \\
    \bottomrule
  \end{tabular}
  }
\end{table}

%% file: tables/clip_model_card.tex
\begin{table}
  \caption{Model card of HPS v2.}
  \label{tab:model_card}
  \centering
  \resizebox{0.4\textwidth}{!}{
  \begin{tabular}{lc}
    \toprule
    Embedding dimension & 1024 \\
    \midrule
    Input image size & 224 \\
    Input patch size & 14 \\
    \# image transformer layers & 32 \\
    Image transformer width & 1280 \\
    Image transformer \# heads & 16 \\
    \midrule
    Context length & 77 \\
    Vocabulary size & 49408 \\
    \# text transformer layers & 24 \\
    Text transformer width & 1024 \\
    Text transformer \# heads & 16 \\
    \bottomrule
  \end{tabular}
  }
\end{table}

%% file: tables/pairwise.tex
\begin{table}
  \caption{Pairwise Preference Prediction Accuracy of HPSv2}
  \label{tab:pairwise eval}
  \centering
  \resizebox{1.0\textwidth}{!}{
  \begin{tabular}{lccccccccc}
    \toprule
     Model & SD v1.4 & SD v2.0 & VQ-Diffusion & LAFITE & GLIDE & CogView2 & VQGAN+CLIP & DALL·E2 & COCO \\ 
     \midrule
     FuseDream & 87.3\% & 92.5\% & 67.8\% & 79.3\% & 90.2\% & 85.5\% & 51.5\% & 90.3\% & 92.0\% \\
     SD v1.4 & - & 72.0\% & 90.3\% & 92.0\% & 96.5\% & 67.5\% & 70.5\% & 78.5\% & 85.0\% \\
     SD v2.0 & - & - & 93.3\% & 97.0\% & 98.8\% & 76.5\% & 83.0\% & 73.8\% & 79.0\% \\
     VQ-Diffusion & - & - & - & 80.8\% & 88.0\% & 90.0\% & 51.5\% & 91.3\% & 93.0\% \\
     LAFITE & - & - & - & - & 82.8\% & 92.0\% & 73.8\% & 92.8\% & 95.0\% \\
     GLIDE & - & - & - & - & - & 96.8\% & 84.8\% & 96.8\% & 96.0\% \\
     CogView2 & - & - & - & - & - & - & 73.0\% & 79.0\% & 84.0\% \\
     VQGAN+CLIP & - & - & - & - & - & - & - & 79.5\% & 92.0\% \\
     DALL·E 2 & - & - & - & - & - & - & - & - & 74.0\% \\
    \bottomrule
  \end{tabular}
}
\end{table}

%% file: tables/ablation_lr.tex
\begin{tabular}{ccc}
    \toprule
    LR/$\times 10^{-6}$ & HPD v2 \\
    \midrule
    7.5 & 82.93 \\
    5.6 & 82.93 \\
    4.6 & 83.13 \\
    3.6 & 83.19 \\
    3.0 & 83.13 \\
    2.5 & 83.13 \\
    2.0 & 83.05 \\
    0.5 & 81.79 \\
    \midrule
    3.3 & 83.27 \\
    \bottomrule
\end{tabular}

%% file: tables/ablation_image.tex
\begin{tabular}{ccc}
    \toprule
    Image Layers & HPD v2 \\
    \midrule
    30 & 83.14 \\
    25 & 83.23 \\
    15 & 82.99 \\
    10 & 82.73 \\
    5 & 82.30 \\
    0 & 78.37 \\
    \midrule
    20 & 83.27 \\
    \bottomrule
\end{tabular}

%% file: tables/ablation_text.tex
\begin{tabular}{ccc}
    \toprule
    Text Layers & HPD v2 \\
    \midrule
    22 & 83.07 \\
    16 & 83.29 \\
    13 & 83.26 \\
    9 & 83.15 \\
    6 & 83.01 \\
    0 & 82.89 \\
    \midrule
    11 & 83.27 \\
    \bottomrule
\end{tabular}

%% file: tables/example_usages.tex
\begin{table}
  \caption{Example usages of HPS v2. }
  \label{tab:example_usages}
  \centering
  \begin{tabular}{lllll}
    \toprule
    Method & Animation & Concept-art & Painting & Photo \\
    \midrule
    Stable Diffusion v1.4~\cite{stable_diffusion} & 27.26 & 26.61 & 26.66 & 27.27 \\
    Retrieval initialization & 27.39 & 26.59 & 26.71 & 27.46 \\
    Human-aligned tuning~\cite{wu2023better} & 27.80 & 27.16 & 27.24 & 27.60 \\
    \bottomrule
  \end{tabular}
\end{table}